\documentclass[sigconf]{aamas}

\usepackage{balance} 
\usepackage{amsmath,amsthm,mathtools}
\usepackage{bbm}
\usepackage{bm}
\usepackage{algorithm}
\usepackage{algpseudocode}
\usepackage{enumitem}
\usepackage{booktabs}
\usepackage{xcolor} 

\newcommand{\R}{\mathbb{R}}

\newcommand{\T}{\top}
\newcommand{\Tau}{\mathcal{T}}

\newtheorem{theorem}{Theorem}
\newtheorem{lemma}[theorem]{Lemma}

\theoremstyle{definition}
\newtheorem{definition}[theorem]{Definition}
\newtheorem{assumption}[theorem]{Assumption}
\newtheorem{remark}[theorem]{Remark}

\doi{PAPW1165}

\makeatletter
\gdef\@copyrightpermission{
  \begin{minipage}{0.2\columnwidth}
   \href{https://creativecommons.org/licenses/by/4.0/}{\includegraphics[width=0.90\textwidth]{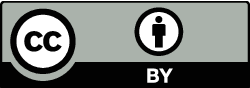}}
  \end{minipage}\hfill
  \begin{minipage}{0.8\columnwidth}
   \href{https://creativecommons.org/licenses/by/4.0/}{This work is licensed under a Creative Commons Attribution International 4.0 License.}
  \end{minipage}
  \vspace{5pt}
}
\makeatother

\setcopyright{ifaamas}
\acmConference[AAMAS '26]{Proc.\@ of the 25th International Conference
on Autonomous Agents and Multiagent Systems (AAMAS 2026)}{May 25 -- 29, 2026}
{Paphos, Cyprus}{C.~Amato, L.~Dennis, V.~Mascardi, J.~Thangarajah (eds.)}
\copyrightyear{2026}
\acmYear{2026}
\acmDOI{}
\acmPrice{}
\acmISBN{}

\title[AAMAS-2026 Formatting Instructions]{MA-SafeDiffuser: Safe Multi-Agent Planning with Diffusion Probabilistic Models}

\author{Kiran Ravish}
\affiliation{
  \institution{IIIT Hyderabad}
  \city{Hyderabad}
  \country{India}}
\email{kiran.ravish@research.iiit.ac.in}

\author{Ankita Kushwaha}
\affiliation{
  \institution{IIIT Hyderabad}
  \city{Hyderabad}
  \country{India}}
\email{ankita.kushwaha@research.iiit.ac.in}

\author{Preeti}
\affiliation{
  \institution{IIIT Hyderabad}
  \city{Hyderabad}
  \country{India}}
\email{preeti.preeti@research.iiit.ac.in}

\author{Pawan Kumar}
\affiliation{
  \institution{IIIT Hyderabad}
  \city{Hyderabad}
  \country{India}}
\email{pawan.kumar@iiit.ac.in}

\begin{abstract}
We propose \emph{MA-SafeDiffuser}, a multi-agent extension of SafeDiffuser that equips diffusion-based trajectory planners with \emph{finite-time diffusion invariance} guarantees for joint-agent safety specifications. Building on the single-agent construction that embeds control barrier function (CBF) constraints into reverse-diffusion updates, we: (i) formalize joint safe sets as intersections of per-agent and pairwise barrier sets; (ii) derive centralized and decentralized (communication-aware) constrained denoising procedures with provable invariance under mild assumptions; (iii) address local-trap and deadlock phenomena via time-varying specifications and liveness CBFs; and (iv) develop a lightweight benchmarking suite including a multi-agent Maze2D domain. Empirically, MA-SafeDiffuser reduces violation counts relative to unconstrained diffusion baselines, while retaining planning quality. We provide algorithms, proofs, and reference implementations.
\end{abstract}


\newcommand{\BibTeX}{\rm B\kern-.05em{\sc i\kern-.025em b}\kern-.08em\TeX}

\begin{document}


\pagestyle{fancy}
\fancyhead{}


\maketitle 


\section{Introduction}

\label{sec:intro}

Diffusion probabilistic models (DDPMs)~\cite{NEURIPS2020_4c5bcfec} were recently proposed as a powerful class of generative models whose iterative denoising dynamics can be repurposed for \emph{trajectory synthesis} and \emph{planning}. Recent exciting work such as Diffuser~\cite{janner2022Diffuser} and subsequent follow-ups on classifier/energy guidance~\cite{DhariwalNichol2021} demonstrate that diffusion models can indeed produce long-horizon behaviors that emulates task-level structure and can also leverage heterogeneous offline experience. When applied casually, however, unconstrained denoising provides no formal safety guarantees: for instance, an optimized sample may graze or penetrate obstacles, violate actuation limits, or collide with other agents.

\paragraph{Extension from diffusion to safe diffusion.}
The SafeDiffuser framework~\cite{SafeDiffuser} addresses this issue in the \emph{single-agent} setting by embedding \emph{Control Barrier Function} (CBF) inequalities~\cite{7782377} into each reverse step of the diffusion process. Conceptually, the unconstrained denoiser proposes an update and a safety projector modifies it to satisfy linearized CBF rows; an execution-time filter then mirrors the same CBF constraints in receding horizon. This yields \emph{finite-time diffusion invariance} for the safe set while retaining the modeling flexibility of diffusion planning. Extensions of CBFs, e.g., exponential CBFs for higher relative degree~\cite{NguyenSreenath2016ECBF} and nonsmooth barrier sets~\cite{NonsmoothCBF} further enlarge the class of safety specifications that can be handled.

\paragraph{Motivation for Multi-agent}
Many real systems such as robot swarms, UAV traffic management, multi-arm manipulation, automated warehouses are inherently \emph{multi-agent}. Safety must hold (i) for each agent with respect to walls, bounds, and actuation limits; (ii) \emph{pairwise} to prevent inter-agent collisions; and sometimes (iii) \emph{globally} for task-level constraints. Classical reactive methods such as artificial potential fields~\cite{Khatib1986PotentialFields}, velocity obstacles and RVO/ORCA~\cite{FioriniShiller1998VO,vanDenBerg2011ORCA}, and model-predictive control with hand-crafted penalties provide strong baselines in practice, but do not yield end-to-end invariance guarantees when coupled with learned generative planners and can deadlock under symmetric interactions. On the learning side, multi-agent RL (e.g., MADDPG~\cite{MADDPG}, MAPPO~\cite{MAPPO}) excels at flexible coordination but typically lacks hard safety certificates and may require extensive online exploration; recent surveys on SafeRL and SafeMARL further highlight these limitations and open challenges in constrained multi-agent settings~\cite{kushwaha2025surveysafeRL}.

\paragraph{Our Contributions.}
We develop \textbf{MA–SafeDiffuser}, a \emph{multi-agent} extension of SafeDiffuser that equips diffusion planning with centralized and decentralized safety enforcement and theoretical guarantees. The key idea is to formalize the \emph{joint safe set} at each waypoint as the intersection of per-agent, pairwise, and (optional) task barriers, and to enforce the corresponding linearized CBF rows during denoising. A two-time formulation (reverse step $j$ over planning waypoints $k$) enables \emph{receding-horizon} enforcement at $k=0$ together with temporal regularization and short multi-$k$ seeding to distribute margin across all $k$.

\paragraph{Key Contributions.}
\begin{itemize}
\item \textbf{Joint safety with diffusion.} A centralized projector that enforces the \emph{conjunction} of per-agent and pairwise CBF rows throughout reverse denoising, with optional task constraints. A receding–horizon execution-time filter mirrors these rows to certify the applied control.
\item \textbf{Finite-time diffusion invariance.} Under standard $C^{1,1}$ regularity, step-size bounds, and feasibility assumptions, we prove that the denoising process is \emph{finite-time diffusion invariant} for the joint safe set, and we provide a discrete-time CBF comparison lemma for reverse-Euler updates.
\item \textbf{Decentralized enforcement with neighbor messaging.} A communication–aware variant in which each agent enforces local self/task rows and \emph{robust} pairwise rows using predicted neighbor states, responsibility shares, and a computable robustness margin that compensates for bounded prediction errors; invariance carries over under clear Lipschitz and step-bound conditions.

\item \textbf{Practical deadlock mitigations.} Time-varying specifications (TVS) that relax early reverse steps, small symmetry-breaking biases in pairwise shares, and progress/liveness barriers toward corridor waypoints to avoid local traps; these mechanisms are orthogonal to the safety proof and improve success rates in practice.

\item \textbf{Evaluation and differentiable enforcement.} 

A light Maze2D benchmark with door-aware geometry and per-step action clamps for fair Safe/Unsafe comparisons; enforcement can be implemented via differentiable QP layers (OptNet, CVXPY-Layers)~\cite{OptNet,DiamondBoyd2016CVXPY} or via parameter–free Dykstra projections onto half-spaces~\cite{Dykstra1983,BauschkeCombettes}, and solves in milliseconds in our experiments.
\end{itemize}
Our approach complements the growing literature on diffusion for decision making and control~\cite{janner2022Diffuser, NEURIPS2020_4c5bcfec, DhariwalNichol2021}, including goal/skill-conditioned and score-guided variants, by \emph{hardening} the generative sampler with CBF certificates. Relative to multi-agent navigation baselines (potential fields, VO/ORCA)~\cite{Khatib1986PotentialFields,FioriniShiller1998VO,vanDenBerg2011ORCA} and MARL~\cite{MADDPG,MAPPO}, MA SafeDiffuser provides a \emph{model-agnostic} safety layer that can be used with hand-crafted (score) denoisers or learned score networks and yields invariance guarantees at both denoising and execution time.

\paragraph{Organization of this paper.}
Section~\ref{sec:problem} formalizes the two-time indexing, barrier classes, and joint safe sets. Section~\ref{sec:method} presents centralized and decentralized projectors and TVS/liveness devices. Section~\ref{sec:theory} establishes diffusion invariance and decentralized robustness. Section~\ref{sec:expts} reports Maze2D results and ablative analyses. Section~\ref{sec:conclusion
-limitations} concludes the paper by summarizing  limitations and future directions (larger agent counts, richer tasks, tighter learned scores).

\section{Problem Formulation and Notation}
\label{sec:problem}

\paragraph{States, trajectories, indices.}
We use the two-time notation of~\cite{SafeDiffuser}: \emph{diffusion time} $j\in\{0,\ldots,N\}$ indexes the reverse steps of the denoiser, the \emph{planning time} $k\in\{0,\ldots,H\}$ indexes the $H{+}1$ waypoints of a trajectory. For $M$ agents, we define agent $a$ has state space $\mathcal{X}_a\subseteq\R^{n_a}$ and we have state $x^{\,j}_{a,k}\in\mathcal{X}_a$ at $(j,k)$. We stack joint states as
\begin{align*}
\bm{x}^{\,j}_k \;=\; \big(x^{\,j}_{1,k},\ldots,x^{\,j}_{M,k}\big)\in \mathcal{X}\;:=\;\mathcal{X}_1\times\cdots\times\mathcal{X}_M,\quad
n\;=\;\sum_{a=1}^M n_a.
\end{align*}
We define a joint trajectory at diffusion step $j$ to be 

$\tau^{\,j}=(\bm{x}^{\,j}_0,\ldots,\bm{x}^{\,j}_H)\in\mathcal{X}^{H+1}$. We use the notation $\Tau:=\mathcal{X}^{H+1}$ for the trajectory space and use bold symbols for joint quantities.

\paragraph{Reverse-time update.}

Let $u^{\,j}\in\R^{n(H+1)}$ denote the stacked reverse-time update at step $j$ (arranged by waypoints and agents). The unconstrained reverse dynamics are formalized via the Euler limit as follows
\begin{equation}
  \dot{\tau}^{\,j}
  \;=\;
  \lim_{\Delta\tau\to 0}\frac{\tau^{\,j}-\tau^{\,j+1}}{\Delta\tau}
  \;=\;
  u^{\,j},
  \: \text{equivalently}, \:
  \tau^{\,j} \;=\; \tau^{\,j+1} + \Delta\tau\,u^{\,j},
  \label{eq:reverse-drift}
\end{equation}
where $u^{\,j}$ is produced by the denoiser’s drift (reference interpolation $+$ temporal smoothing) and this is subsequently \emph{projected} to enforce safety constraints~\cite{SafeDiffuser,janner2022Diffuser}.

\paragraph{Safety specifications.}
At each waypoint $k$, we have three classes of (Clarke-regular) barrier functions as follows:
\begin{itemize}
\item \emph{Per-agent barriers} $b^{\mathrm{self}}_{a,k}:\mathcal{X}_a\to\R$ with safe super-level sets $\{x_{a,k}: b^{\mathrm{self}}_{a,k}(x_{a,k})\ge 0\}$,
\item \emph{Pairwise barriers} $b^{\mathrm{pair}}_{ab,k}:\mathcal{X}_a\times\mathcal{X}_b\to\R$ ($a<b$) with safe sets $\{(x_{a,k},x_{b,k}): b^{\mathrm{pair}}_{ab,k}\ge 0\}$,
\item (optional) \emph{Task or global barriers} $b^{\mathrm{task}}_{k}:\mathcal{X}\to\R$ with safe set defined as $\{\bm{x}_k: b^{\mathrm{task}}_{k}(\bm{x}_k)\ge 0\}$.
\end{itemize}
The \emph{joint waypoint-wise safe set} now can be defined as follows
\begin{align}
  \mathcal{C}_k
  \;:=\;
  &\bigcap_{a=1}^M \bigl\{x_{a,k}: b^{\mathrm{self}}_{a,k}(x_{a,k})\ge 0\bigr\} \nonumber \\
  &\;\cap\;
  \bigcap_{a<b}\bigl\{(x_{a,k},x_{b,k}): b^{\mathrm{pair}}_{ab,k}(x_{a,k},x_{b,k})\ge 0\bigr\} \nonumber \\ 
  &\;\cap\;
  \bigl\{\bm{x}_k: b^{\mathrm{task}}_{k}(\bm{x}_k)\ge 0\bigr\}.
  \label{eq:joint-safe-set}
\end{align}
We can now extend this safe waypoint notion to safe trajectories: a trajectory $\tau=(\bm{x}_0,\ldots,\bm{x}_H)\in\Tau$ is \emph{safe} iff $\bm{x}_k\in\mathcal{C}_k$ for all $k$. The \emph{trajectory-level safe set} is then written as follows
\[
\mathcal{C} \;:=\; \bigl\{\tau\in\Tau:\ \bm{x}_k\in\mathcal{C}_k\ \text{for all }k=0,\ldots,H\bigr\}.
\]

\begin{definition}[Multi-agent finite-time diffusion invariance]
\label{def:ftdi}
The reverse-time denoising process is \emph{finite-time diffusion invariant} for $\mathcal{C}$ if there exists an index $i\in\{0,\ldots,N\}$ such that, for every $j\le i$, the denoised trajectory $\tau^{\,j}$ is safe: $\tau^{\,j}\in\mathcal{C}$, i.e., $\bm{x}^{\,j}_k\in\mathcal{C}_k$ for all $k$.
\end{definition}

\begin{remark}[Enforcement index and receding horizon]

When it comes to implementing CBFs, we usually apply the linearized constraints at the very first point in time, or $k=0,$ which is referred to as the receding horizon, and then rely on temporal regularization of the reverse model and a very short period of multi-$k$ enforcement to get a safety net that covers the rest of the planning horizon. See \S\ref{sec:method} and \S\ref{sec:theory}).
\end{remark}


\section{Method: MA-SafeDiffuser}
\label{sec:method}

In our multi-agent extension, we enforce safety by \emph{projecting} the unconstrained reverse-diffusion drift onto linearized
CBF half-spaces at a designated enforcement index. By default we choose $k=0,$ as receding horizon at \emph{every} reverse step $j$. The same filter is applied at execution time to the instantaneous control. In our proposal we have the following:
(i) a \emph{central} projector that is responsible for enforcing all per-agent, pairwise, and task rows jointly (see \S\ref{sec:centralized}); (ii) a \emph{decentralized} variant with neighbor messaging and robust margins (see \S\ref{sec:decentralized}); and (iii) a set of time-varying specifications and liveness mechanisms that avoids deadlock (see \S\ref{sec:liveness}). 
The regularization of the reverse model and/or a brief period of enforcing the model at different values of $k$ send the margin to all values of $k,$ when the constraints are only applied at $k=0$ (see cf.\ \S\ref{sec:theory}).

\section{Algorithms}
\label{sec:algorithms}

\begin{algorithm}[t!]
\caption{\textsc{MA-SafeDiffuser:} Centralized projector (with one reverse step at $j$). Here \(\gamma:[0,N]\to\mathbb{R}_{\ge 0}\) is the TVS \emph{relaxation schedule}-a nonincreasing function of the reverse index \(j\) (with \(\gamma(0)=0\)) that temporarily offsets the barrier and whose (discrete) diffusion-time derivative is denoted \(\dot\gamma(j)\).}
\label{alg:central}
\begin{algorithmic}[1]
\Require Previous trajectory $\tau^{\,j+1}\!=\!(\bm{x}^{\,j+1}_0,\ldots,\bm{x}^{\,j+1}_H)$; unconstrained reverse drift $u^{\,j}_0$ (from \eqref{eq:reverse-drift}); barrier rows $\{b_\ell\}$ at enforcement index $k=0$; variant $\in\{\mathrm{RoS},\mathrm{ReS},\mathrm{TVS}\}$; step $\Delta\tau$; step bound $u_{\max}$; weights $(\lambda,\omega^{\,j})$
\Ensure Projected update $u^{\,j}$ and next trajectory $\tau^{\,j}$
\State \textbf{Linearize barriers at $k=0$:} for each active row $b_\ell$ (self/pair/task) form gradient
$\nabla b_\ell(\bm{x}^{\,j}_0)$ and right-hand side $h_\ell^{\,j}$:
\Statex \textbf{RoS/ReS:} $h_\ell^{\,j}\gets \alpha_\ell(b_\ell(\bm{x}^{\,j}_0))$;\\
\textbf{TVS:} $h_\ell^{\,j}\gets \alpha_\ell(b_\ell(\bm{x}^{\,j}_0)-\gamma(j))-\dot\gamma(j)$. 

\State \textbf{Build convex program} in decision variables $(u^{\,j},r^{\,j})$ as follows
\begin{align*}
&\min~~\big\|u^{\,j}-u^{\,j}_0\big\|_2^2+\lambda\|r^{\,j}\|_2^2 \\ 
&\text{s.t.}\quad
\nabla b_\ell(\bm{x}^{\,j}_0)^{\!\T}u^{\,j} \;\ge\; -\,\omega_\ell^{\,j} r_\ell^{\,j}+h_\ell^{\,j},~~
\|u^{\,j}_a\|_2\le u_{\max}~\forall a.
\end{align*}
\State \textbf{Solve} (QP or Dykstra on half–spaces) to obtain $(u^{\,j},r^{\,j})$; set $r^{\,j}\!=\!0$ in RoS.
\State \textbf{Update reverse step:} $\tau^{\,j}\leftarrow \tau^{\,j+1} + \Delta\tau\,u^{\,j}$.
\State \textbf{(Optional all-$k$ enforcement)} If required (cf.\ \S\ref{sec:theory}), repeat 1–5 with rows at all $k$ for a fixed number of reverse steps.
\State \textbf{return} $\tau^{\,j}$.
\end{algorithmic}
\end{algorithm}

\subsection{Centralized joint enforcement}
\label{sec:centralized}

For the reverse step $j$ we let $u^{\,j}_0\in\R^{n(H+1)}$ denote the stacked unconstrained drift (cf.\ \eqref{eq:reverse-drift}). 
We solve a minimum-deviation objective at index $k=0,$ by linearizing the active barriers as follows.

\begin{align}
\min_{u^{\,j},\,r^{\,j}}~~&
\big\|u^{\,j} - u^{\,j}_0\big\|_2^2 \;+\; \lambda\,\big\|r^{\,j}\big\|_2^2
\qquad\text{s.t.}\nonumber\\[2pt]
\text{(self)}~~&
\nabla b^{\mathrm{self}}_{a,0}\!\big(x^{\,j}_{a,0}\big)^{\!\T} u^{\,j}_{a}
\;+\; \alpha\big(b^{\mathrm{self}}_{a,0}(x^{\,j}_{a,0})\big)
~\ge~ -\,\omega^{\,j}_{a}\, r^{\,j}_{a},
\label{eq:self-cbf} \\ 
&\forall a\in\{1,\ldots,M\}, \nonumber \\[1pt]
\text{(pair)}~~&
\nabla_{x_a} b^{\mathrm{pair}}_{ab,0}\!\big(x^{\,j}_{a,0},x^{\,j}_{b,0}\big)^{\!\T} u^{\,j}_{a}
\;+\;
\nabla_{x_b} b^{\mathrm{pair}}_{ab,0}\!\big(x^{\,j}_{a,0},x^{\,j}_{b,0}\big)^{\!\T} u^{\,j}_{b} \nonumber \\ 
&\;+\; \alpha\big(b^{\mathrm{pair}}_{ab,0}(\cdot)\big)
~\ge~ -\,\omega^{\,j}_{ab}\, r^{\,j}_{ab}, \quad \forall a<b,
\label{eq:pair-cbf}\\[1pt]
\text{(task)}~~&
\nabla b^{\mathrm{task}}_{0}\!\big(\bm{x}^{\,j}_0\big)^{\!\T} u^{\,j}
\;+\; \alpha\big(b^{\mathrm{task}}_{0}(\bm{x}^{\,j}_0)\big)
~\ge~ -\,\omega^{\,j}_{0}\, r^{\,j}_{0}
\label{eq:task-cbf}
\end{align}

here $u^{\,j}=(u^{\,j}_1,\ldots,u^{\,j}_M)$, $\alpha$ is an extended class-$\mathcal{K}$ function,
and $r^{\,j}$ collects relaxation variables. The variants are:
\begin{itemize}
\item \textbf{RoS} (robust-safe): here we consider hard enforcement with $r^{\,j}\equiv 0$ and all $\omega^{\,j}\equiv 0$.
\item \textbf{ReS} (relaxed-safe): here we allow $r^{\,j}\neq 0$ with diffusion–time weights $\omega^{\,j}\downarrow 0$
      to recover feasibility; feasibility gaps vanish as $j$ decreases.
\item \textbf{TVS} (time-varying-safe): here we replace each $b$ by $b-\gamma(j)$ and augment the RHS by $-\dot\gamma(j)$
      (the linearization is applied to $b-\gamma$).
\end{itemize}

During numerical simulations, imposing step bounds (e.g., $\|u^{\,j}_{a}\|_2 \le u_{\max}$ for each agent block) is useful for controlling truncation errors and ensuring numerical stability. 
The resulting QP can be solved using any sparse convex QP solver (e.g., OSQP or CVXPY-Layers~\cite{DiamondBoyd2016CVXPY}). 
When all constraints are linearized half-spaces, one may alternatively employ Dykstra-type projection methods~\cite{Dykstra1983,BauschkeCombettes} (see \S\ref{sec:execution}).

\paragraph{All-$k$ enforcement (optional).}
In case, the denoiser lacks a temporal contraction, we may enforce \eqref{eq:self-cbf}–\eqref{eq:task-cbf}
for all $k\in\{0,\ldots,H\}$ during the first $J$ reverse steps; this will ensure a safety margin at every waypoint
that persists under the subsequent $k=0$-only enforcement (cf.\ \ref{thm:centralized-invariance}), however, at the expense of more possibly more compute.

\subsection{Decentralized enforcement with neighbor messaging}
\label{sec:decentralized}
In an environment with multiple agents and limited long range communication, we may consider a decentralized variant of our method. To this end, each agent $a$ solves a \emph{local} projector using \emph{predicted} neighbor states denoted by $\widehat{x}^{\,j}_{b,0}.$ These predicted neighbor states are obtained from the previous communication round, for example. In the decentralized setting, we need to localize the barriers. For pairwise barriers, we linearize locally at $(x^{\,j}_{a,0},\widehat{x}^{\,j}_{b,0})$ and we enforce a \emph{robust} inequality with
\emph{responsibility sharing} $\rho_{ab},\rho_{ba}\in(0,1)$, $\rho_{ab}{+}\rho_{ba}=1$:
\[
\nabla_{x_a} b^{\mathrm{pair}}_{ab,0}\!\big(x^{\,j}_{a,0},\widehat{x}^{\,j}_{b,0}\big)^{\!\T} u^{\,j}_{a}
\;+\; \rho_{ab}\,\alpha\!\big(b^{\mathrm{pair}}_{ab,0}(x^{\,j}_{a,0},\widehat{x}^{\,j}_{b,0})\big)
~\ge~ m_{\mathrm{rob}},
\]
(and similarly for agent $b$). With one to three Jacobi and  ADMM-style rounds~\cite{ADMM} per diffusion step,
messages are updated and the local projections are recomputed. Under prediction error
$\|\widehat{x}^{\,j}_{b,0}-x^{\,j}_{b,0}\|\le \varepsilon$, Lipschitz constants $L_b$ (barrier value),
$L_{\nabla}$ (barrier gradients), $L_\alpha$ (class-$\mathcal{K}$), and per–step bounds $\|u^{\,j}_{a}\|\le u_{\max}$,
it suffices (see Theorem \ref{thm:decentralized-invariance}) to pick
\begin{equation}
\label{eq:mrob-bound}
m_{\mathrm{rob}} \;\ge\; \tfrac{1}{2}\,\Big(2\,L_{\nabla}\,u_{\max} \;+\; L_\alpha\,L_b\Big)\,\varepsilon,
\end{equation}
to guarantee that the \emph{true} centralized pairwise inequalities (with $(x^{\,j}_{a,0},x^{\,j}_{b,0})$)
are satisfied (see Theorem \ref{thm:decentralized-invariance} for a detailed proof). We note here that self and task barriers do not depend on
predictions and they can be enforced locally similar to the centralized case.

\subsection{Execution-time safety filter}
\label{sec:execution}

On completion of the reverse steps at index $j$, we get a first waypoint proposal, namely,  $x^{\,j}_{0}\mapsto x^{\,j}_{1}$.
Before execution at task time $t$, we solve the \emph{instantaneous} CBF projection for $u_t=x_{t+1}-x_{t}$ with the
same rows (now evaluated at $\bm{x}_t$) and alongwith explicit step bounds, e.g., $\|u_t\|_2\le u_{\max}$ (or for example,
$\|u_t\|_\infty\le u_{\max}$). As mentioned before, this receding-horizon filter prevents rare large steps from creating violations and
mirrors SafeDiffuser’s deployment~\cite{SafeDiffuser}. When constraints are linearized half-spaces, Dykstra’s algorithm
provides a parameter-free projection onto the intersection of half–spaces~\cite{Dykstra1983,BauschkeCombettes}; otherwise, we may prefer to solve a small convex QP problem. The execution-time filter is essential to preserve invariance for the applied control (cf.\
\ref{thm:centralized-invariance}).

\subsection{Avoiding traps and deadlock}
\label{sec:liveness}
Despite the safety constraints using control barriers, we may encounter deadlocks, and local traps. To this end, we propose to employ three practical devices to reduce deadlock and local traps without sacrificing (safety) invariance:
\begin{enumerate}
\item \textbf{Time-varying specifications (TVS).} Use $b-\gamma(j)$ with $\gamma$ nonincreasing in reverse time and $\gamma(0)=0$; constraints are relaxed early and tightened near $j=0.$
\item \textbf{Symmetry breaking.} We add small bias offsets or priorities in pairwise shares $(\rho_{ab},\rho_{ba})$
      to avoid stalemates in symmetric configurations. For example, $\rho_{ab} > \rho_{ba}$ will avoid the symmetry. 
\item \textbf{Liveness/progress CBFs.} We include auxiliary barriers that encode progress toward intermediate subgoals for example, corridor waypoints, ensuring nonzero ``driving'' terms when pure safety is noninformative.
\end{enumerate}
As we show later, these mechanisms are orthogonal to the core safety projector and are compatible with the invariance proofs we derive in Section \ref{sec:theory}.

\section{Theoretical Results}
\label{sec:theory}

We prove finite-time diffusion invariance under standard smoothness and feasibility assumptions, and we also provide a decentralized robustness guarantee.

\begin{assumption}[Regularity and feasibility]
\label{asm:regularity}
All active barriers are assumed to be $b_\ell$ are $C^{1,1}$ on an open set $\mathcal{U}\supset\mathcal{C}$, i.e., $b_\ell$ are
continuously differentiable and we assume that their gradients are locally Lipschitz on $\mathcal{U}$; moreover, for each barrier, there
exists an extended class-$\mathcal{K}$ function $\alpha_\ell$.
At every reverse step $j$, the projector~\eqref{eq:self-cbf}-\eqref{eq:task-cbf} (RoS) is feasible or admits
relaxations (ReS/TVS) with $\omega^{\,j}\!\downarrow 0$. The unconstrained drift $u^{\,j}_0$ is assumed to be locally Lipschitz
in $\tau^{\,j}$ and that it satisfies per-step bounds $\|u^{\,j}_{a}\|\le u_{\max}$.
\end{assumption}

\begin{lemma}[Barrier conjunction]
\label{lem:conjunction}
Let $\mathcal{C}=\bigcap_{\ell=1}^L\{z:\,b_\ell(z)\ge 0\}$ and consider $\dot{z}=u(z)$.
If for all $\ell$ and all $z$,
\,$\nabla b_\ell(z)^\T u(z)+\alpha_\ell(b_\ell(z))\ge 0$,
then $\mathcal{C}$ is forward invariant, that is, we have  $z(0)\in\mathcal{C}\Rightarrow z(t)\in\mathcal{C}$ for all $t\ge 0$.
\end{lemma}
\begin{proof}[Proof Sketch]
We know that each single set $S_\ell=\{b_\ell\ge 0\}$ is forward invariant by the CBF condition; as we know that at boundary points
$\{b_\ell=0\}$ the field is tangent–inward~\cite{7782377,NonsmoothCBF,Nagumo}.
For smooth inequality sets, the tangent cone of the intersection equals the intersection of tangent cones,
so $u(z)\in T_{S_\ell}(z)$ for all active $\ell$ implies $u(z)\in T_{\mathcal{C}}(z)$, and invariance will follow from
Nagumo’s theorem \cite{Nagumo}. The same conclusion will hold for locally Lipschitz $b_\ell$ when using Clarke generalized gradients. More detail can be found in supplementary. 
\end{proof}

\begin{algorithm}[t!]
\caption{\textsc{MA-SafeDiffuser-DEC:} Decentralized projector (one reverse step at $j$)}
\label{alg:decentral}
\begin{algorithmic}[1]
\Require Neighbor graph $\mathcal{G}$; last–round neighbor predictions $\widehat{x}^{\,j}_{b,0}$; robust margin $m_{\rm rob}$; shares $\rho_{ab},\rho_{ba}\!\in\!(0,1)$, $\rho_{ab}{+}\rho_{ba}\!=\!1$; per–agent bound $u_{\max}$
\Ensure Local updates $\{u^{\,j}_a\}_{a=1}^M$ and aggregated $u^{\,j}$
\For{$a\in\{1,\ldots,M\}$ \textbf{in parallel}}
  \State \textbf{Self/task rows:} build $\nabla b^{\mathrm{self}}_{a,0}(x^{\,j}_{a,0})$ and (if present) global rows as in Alg.~\ref{alg:central}.
  \State \textbf{Pairwise rows (robust):} for each $b\in\mathcal{N}(a)$, linearize at $(x^{\,j}_{a,0},\widehat{x}^{\,j}_{b,0})$ and enforce
  \[
  \nabla_{x_a} b^{\mathrm{pair}}_{ab,0}(x^{\,j}_{a,0},\widehat{x}^{\,j}_{b,0})^{\!\T} u^{\,j}_a
  \;+\; \rho_{ab}\,\alpha\!\big(b^{\mathrm{pair}}_{ab,0}(x^{\,j}_{a,0},\widehat{x}^{\,j}_{b,0})\big)
  \;\ge\; m_{\rm rob}.
  \]
  \State \textbf{Local solve:} minimize $\|u^{\,j}_a-u^{\,j}_{0,a}\|_2^2$ subject to the above and $\|u^{\,j}_a\|_2\le u_{\max}$; broadcast updated $x^{\,j}_{a,0}$ to $\mathcal{N}(a)$.
\EndFor
\State \textbf{Repeat} the local step for $R\in\{1,2,3\}$ message–passing rounds.
\State \textbf{Aggregate} $u^{\,j}\leftarrow (u^{\,j}_1,\ldots,u^{\,j}_M)$ and \textbf{update} $\tau^{\,j}\leftarrow \tau^{\,j+1}+\Delta\tau\,u^{\,j}$.
\State \textbf{return} $\tau^{\,j}$.
\end{algorithmic}
\end{algorithm}

\paragraph{A discrete-time CBF lemma (reverse time).}
We use the following Euler-discretized comparison, which basically follows from $C^{1,1}$ regularity.

\begin{lemma}[Discrete-time CBF monotonicity]
\label{lem:disc-cbf}
Let $b\in C^{1,1}$ in a neighborhood of $\mathcal{C}$ with locally Lipschitz gradient constant $L_{\nabla b}$ and
let $\alpha$ be extended class-$\mathcal{K}$. Suppose $u$ satisfies
$\nabla b(x)^\T u + \alpha(b(x))\ge 0$ and consider the reverse-time Euler step
$x^j = x^{j+1} + \Delta\tau\,u^j$ with $\|u^j\|\le u_{\max}$. Then for $\Delta\tau\le \bar\Delta :=
\min\{1,\,\alpha'(0)/(2L_{\nabla b}u_{\max})\}$ there exists $c>0$ such that
\[
b(x^j) \;\ge\; b(x^{j+1}) \;-\; \Delta\tau\,\alpha \big(b(x^j)\big) \;-\; c\,\Delta\tau^2
\]
In particular, we have that if $b(x^{j+1})\ge 0$ then $b(x^j)\ge 0$ \\
for all $\Delta\tau\le\bar\Delta$.
\end{lemma}

\begin{proof}[Proof Sketch]
By $C^{1,1}$, 
$$b(x^j)=b(x^{j+1})+\nabla b(x^j)^\T(x^j{-}x^{j+1})+O(\|x^j{-}x^{j+1}\|^2)$$ with the remainder
bounded by $L_{\nabla b}\Delta\tau^2\|u^j\|^2$. We substitute $x^j{-}\\
x^{j+1}=\Delta\tau\,u^j$ and then use $\nabla b(x^j)^\T u^j\ge -\alpha(b(x^j))$. The step-size condition ensures the implicit inequality preserves
nonnegativity when $b(x^{j+1})\ge 0$ (discrete comparison).
\end{proof}

\begin{remark}[Finite-time reach under RoS/TVS]
\label{rem:finite-time}
If $b(x^N)<0$ (RoS), one can choose $\alpha$ with finite-time property
$\int_{-a}^{0}\frac{ds}{\alpha(s)}<\infty$ on $[-a,0]$ and $\Delta\tau$ small so that the $O(\Delta\tau^2)$
truncation is dominated by $\Delta\tau\,\alpha(\cdot)$; then the discrete recursion crosses into $b\ge 0$ in finitely
many steps. Under TVS with $\tilde b:=b-\gamma(j)$, $\gamma(0)=0$, the same argument would apply to $\tilde b$, giving
$\tilde b(x^0)\ge 0$.
\end{remark}

\begin{theorem}[Centralized multi–agent finite–time diffusion invariance]
\label{thm:centralized-invariance}
Under Assumption \ref{asm:regularity}, if at every reverse step $j$ the centralized projector enforces
\eqref{eq:self-cbf}-\eqref{eq:task-cbf} (RoS or TVS) for all per-agent, pairwise, and task barriers at the
enforcement index, which by default is $k=0$, then the denoising process is finite-time diffusion invariant for
$\mathcal{C}$ with probability $1$ (w.r.t. model noise). Moreover, the execution-time safety filter preserves
invariance for the applied control $u_t$.
\end{theorem}

\begin{proof}[Sketch of Proof]
We fix any barrier $b_\ell$ at $k=0.$ By Lemma \ref{lem:disc-cbf}, the enforced inequality implies no boundary
crossing in reverse time (considering small $\Delta\tau$), so $b_\ell\big(\bm{x}^{\,j}_0\big)\ge 0$ for all $j\le i$
for some $i\le N$ (note that finite-time reach holds by Remark \ref{rem:finite-time}). By Lemma \ref{lem:conjunction}, the conjunction
over all active rows is forward invariant at $k=0.$ If all $k$ are enforced, we are basically done. Otherwise, we either
(i) enforce all $k$ for finitely many reverse steps to seed a margin everywhere, or (ii) we rely on temporal
regularization of the reverse model to propagate a margin from $k=0$ to all $k$ (both are standard in the baseline 
SafeDiffuser~\cite{SafeDiffuser} paper). The “probability~1” qualifier  here is conditional on the event that each QP
(or projection) is feasible; by Assumption \ref{asm:regularity} this event has probability 1 since the projection is a
deterministic measurable map of the model noise. The execution–time CBF filter then basically satisfies
$\nabla b_\ell(\bm{x}_t)^\T u_t + \alpha_\ell(b_\ell(\bm{x}_t))\ge 0$, hence preserves invariance per a
discrete-time CBF argument identical to Lemma \ref{lem:disc-cbf}. More detail can be found in appendix.
\end{proof}

\begin{theorem}[Decentralized invariance under bounded error]
\label{thm:decentralized-invariance}
Assume neighbor messages yield predictions $\widehat{x}^{\,j}_{b,0}$ with bounded error
$\|\widehat{x}^{\,j}_{b,0}-x^{\,j}_{b,0}\|\le\varepsilon$. Suppose each agent enforces its local pairwise CBF
with a robust margin $m_{\rm rob}>0$ and responsibility shares $\rho_{ab},\rho_{ba}\in(0,1)$, $\rho_{ab}{+}\rho_{ba}=1$.
Let $L_b$ and $L_{\nabla}$ be local Lipschitz constants for $b_{ab}$ and its gradients, $L_\alpha$ for $\alpha$, and
assume per-step bounds $\|u^{\,j}_{a}\|\le u_{\max}$. If we have
\begin{equation}
\label{eq:mrob-explicit}
m_{\rm rob} \;\ge\; \tfrac{1}{2}\,\Big(2\,L_{\nabla}\,u_{\max} \;+\; L_\alpha\,L_b\Big)\,\varepsilon
\end{equation}
then the decentralized projected updates satisfy the \emph{true} centralized pairwise inequalities. Consequently,
the conclusions of \ref{thm:centralized-invariance} hold.
\end{theorem}
\begin{proof}[Sketch of Proof]
By Summing the two local inequalities (one at agent $a$, one at agent $b$) built with predicted neighbor states,
and compare them to the centralized inequality evaluated at the \emph{true} joint state. Use Lipschitz bounds
on barrier values and gradients and the per-step bound to show the centralized left-hand side exceeds the
predicted one minus the error term $(2L_{\nabla}u_{\max}+L_\alpha L_b)\varepsilon$. Choosing
$m_{\rm rob}$ as in \eqref{eq:mrob-explicit} makes the sum nonnegative, hence the true centralized pairwise
inequality holds. Taking the finite conjunction over all pairs and combining with self/task rows (which do not
require predictions) gives the desired centralized feasibility, so \ref{thm:centralized-invariance} applies.
\end{proof}

\begin{remark}[Complexity and solvers]
The centralized QP has dimension $(H{+}1)n$ and $O(M^2)$ sparse rows from pairwise constraints. First-order
operator-splitting solvers (e.g., OSQP~\cite{OSQP}) and differentiable layers (OptNet~\cite{OptNet},
CVXPY-Layers \cite{DiamondBoyd2016CVXPY}) are suitable; when constraints are linearized half-spaces, Dykstra-type projections
offer a parameter-free alternative~\cite{Dykstra1983,BauschkeCombettes}.
\end{remark}

\begin{algorithm}[t!]
\caption{Execution-time safety filter (receding horizon at task time $t$)}
\label{alg:exec}
\begin{algorithmic}[1]
\Require Current state $\bm{x}_t$; proposal $\bm{x}_{t+1}^{\mathrm{prop}}$; bound $u_{\max}$
\Ensure Safe control $u_t$ and applied next state $\bm{x}_{t+1}$
\State $u_t^{\,0}\leftarrow \bm{x}_{t+1}^{\mathrm{prop}}-\bm{x}_t$
\State Project $u_t^{\,0}$ onto the intersection of instantaneous rows
$\nabla b_\ell(\bm{x}_t)^{\!\T}u_t \ge \alpha_\ell\!\big(b_\ell(\bm{x}_t)\big)$ and $\|u_t\|_2\le u_{\max}$ (QP or Dykstra)
\State $\bm{x}_{t+1}\leftarrow \bm{x}_t + u_t$;\quad \textbf{return} $(u_t,\bm{x}_{t+1})$
\end{algorithmic}
\end{algorithm}


\section{Numerical Experiments}
\label{sec:expts}

We evaluate \textbf{MA-SafeDiffuser} in  \textbf{Maze2D} domain to validate multi-agent safety and study the
effect of reward shaping and fairness clamps. All experiments were done on RTX 4090 GPUs using python and standard pytorch libraries like pytorch and numpy in double precision arithmetic.

\subsection{Maze2D (Multi-Agent) setup}
\label{sec:exp-maze2d}

\begin{figure*}[t]
    \centering
    \includegraphics[width=0.9\textwidth]{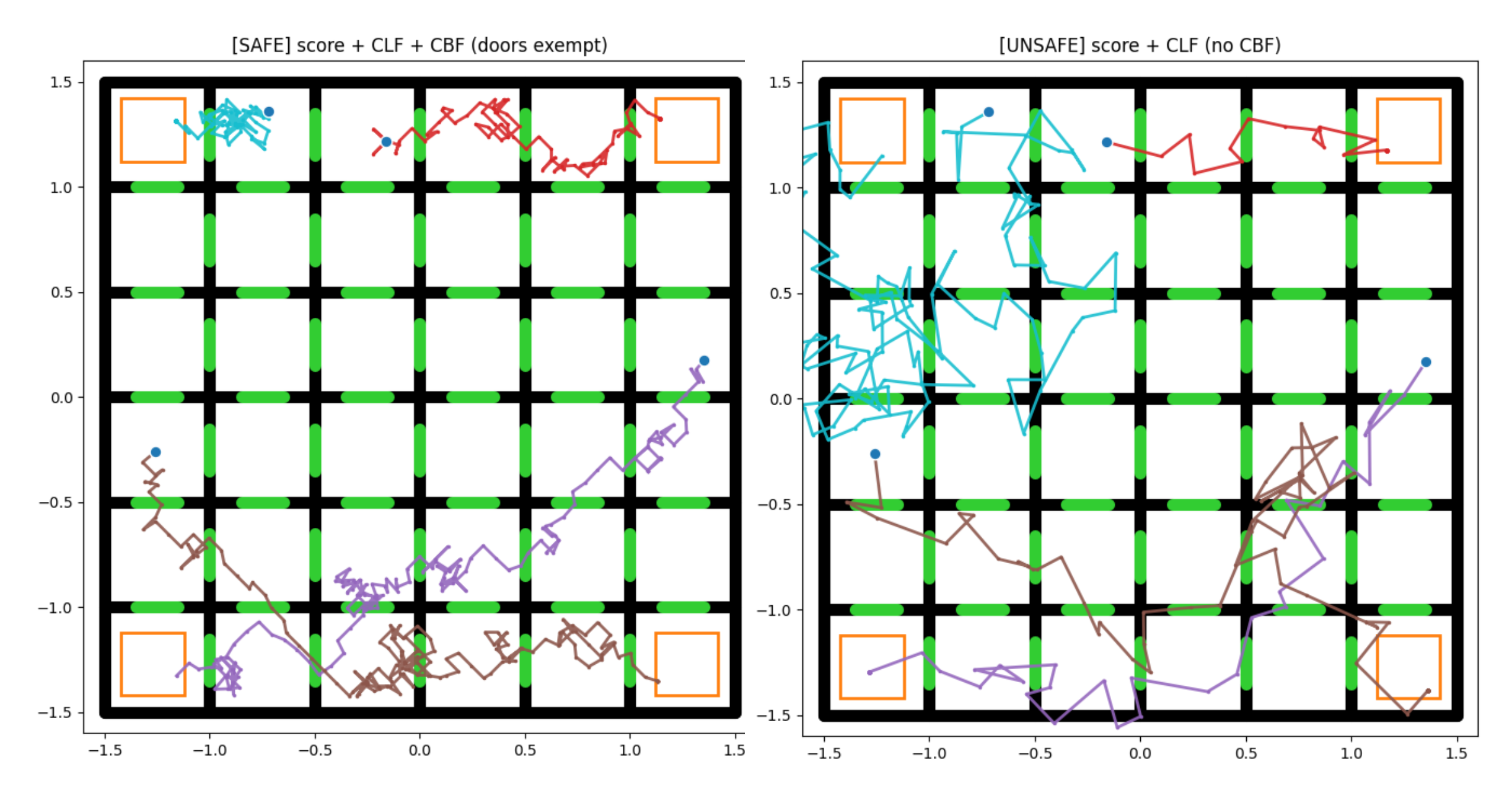}
    \caption{
Qualitative rollouts of the learned-score variant of \textbf{MA-SafeDiffuser}
with an MLP denoiser (4 agents, planning horizon $H{=}28$, diffusion steps $N{=}40$,
and $T{=}2000$ sampling steps; trained on 1500 trajectories for 1000 epochs).
(\textbf{Left}) Safe planner with score-based generation combined with CLF (Control Lyapunov Function) and CBF (Control Barrier Function)
projection, yielding $V_{\text{safe}}{=}62$.
(\textbf{Right}) Unsafe planner without CBF enforcement, yielding
$V_{\text{unsafe}}{=}125$.
The safe variant produces smooth, wall-aware trajectories and avoids collisions,
whereas the unsafe variant exhibits frequent constraint violations and unstable
motion near obstacles.
}

    \label{fig:qualitative_maze}
    \Description{
Qualitative comparison of multi-agent trajectories in a maze environment.
The left panel shows safe trajectories generated using CLF and CBF projection,
where agents avoid walls and inter-agent collisions. The right panel shows unsafe
trajectories without barrier function enforcement, including wall penetration,
collisions, and unstable behavior near obstacles.
}
\end{figure*}

\paragraph{Environment.}
The Maze2D domain is a continuous planar environment bounded within the
axis–aligned box $\mathcal{B} = \{x \in \mathbb{R}^2 : (-1.5, -1.5) \le x \le (1.5, 1.5)\}$.
The interior of $\mathcal{B}$ is partitioned into a $6 \times 6$ grid of square rooms,
each separated by solid wall segments except for small open passages (doors) centered
on the walls. 
Each wall is represented as a line segment $w_\ell = [a_\ell, b_\ell]$, while each door
is modeled as a short open gap $(p_0, p_1)$ on the same line. 
Walls are impassable obstacles, whereas doors define narrow traversable corridors
through which agents can pass. 
In the Figures \ref{fig:qualitative_maze}, \ref{fig:Centralised}, walls are shown in \textbf{black}, door openings in \textbf{green}, and
goal regions as \textbf{orange boxes}. 
Agents are depicted as colored points that begin near random valid start positions
(blue markers) and must navigate to their respective goal boxes while avoiding wall
and inter-agent collisions. We evaluate configurations with 
$M \in \{2,4,8\}$ agents.
The kinematic update rule follows $x_{i,t+1} = x_{i,t} + u_{i,t}$ with bounded control
$\|u_{i,t}\|_2 \le u_{\max}$, and all motion takes place continuously in $\mathbb{R}^2$.

\paragraph{Safety constraints.}
(i) \textbf{Box invariance}: $x_{i,t}\in\mathcal{B}$; \;
(ii) \textbf{Wall clearance}: $d(x_{i,t},w_\ell)\ge \rho_{\rm agent}{+}\rho_{\rm wall}$ with
$(\rho_{\rm agent},\rho_{\rm wall})=(0.05,0.06)$; \;
(iii) \textbf{No wall crossing}: $[x_{i,t},x_{i,t+1}]\cap w_\ell=\varnothing$; \;
(iv) \textbf{Pairwise separation}: $\|x_{i,t}-x_{j,t}\|_2\ge \rho_{\rm pair}=0.25$ for $i\ne j$.
The joint waypoint–wise safe set $\mathcal{C}_k$ is given by \eqref{eq:joint-safe-set}.

\paragraph{Planner.}
At each reverse step, MA-SafeDiffuser performs a score-guided denoising update followed by a projection onto the linearized safety set via Dykstra iterations. Both centralized and decentralized projection modes are supported: the centralized variant enforces all inter-agent constraints jointly, while the decentralized variant performs per-agent local projections using neighbor information within a communication radius of 0.9.
The reference trajectory is generated via door-aware interpolation between start and goal waypoints.

\paragraph{Reward shaping and metrics.}

We measure three metrics per episode:
(i) total violation count $V=\sum_t N_t^{\rm viol}$ (primary safety metric);
(ii) total return $R=\sum_t r_t$ based in goal progress, success bonus (+15) and step penalty (-0.002); (iii) success rate S= fraction of agents that reach goals before horizon T=700. 
To equalize control effort across planners, both Safe and Unsafe variants use identical action clamp 
$|| u_t||_2 \leq u_{max} = 0.085$.

We use two shaping variants:
\begin{itemize}
\item \textit{Progress-dominant shaping} rewards forward motion toward the goal and completion bonuses with mild step costs.
\item \textit{Safety shaping} augments with small penalties for wall proximity and intersection events, enabling finer differentiation between near-miss and hard violation.
\end{itemize}

\paragraph{Baselines and fairness.}
\textsc{Unsafe} = unconstrained denoising (no safety filter). \;
\textsc{Safe} = MA–SafeDiffuser diffusion planner with CBF-style projection.
\textsc{Centralized vs. Decentralized} = comparison between full-state projection and limited-neighbor projection under identical seeds and random noise.
All runs use shared random seeds to ensure fair comparison under equivalent stochasticity.

\paragraph{Protocol and hyperparameters.}
Each configuration is run for three random seeds, across diffusion horizon $H \in \{20,28,36\}$, $N \in \{10,16,24\}$. Planner hyperparameters are fixed as $\beta =0.70$, smoothing $\lambda_{smooth}=0.35$, $\alpha_k = 3.0$, $\text{projection iterations} = 80$, $\Delta \tau = 0.06$. We evaluate both Safe and Unsafe planners over a 700-step horizon, resulting in $54$ configurations across $3$ seeds in total.

\paragraph{Violation counting.} 
We follow a strict safety accounting scheme in all Maze2D experiments. 
At each timestep, the environment monitors three types of safety infractions: 
(i) \textit{soft proximity violations} when an agent enters the near-wall buffer region 
$d(x_i, w_\ell) < \rho_{\text{agent}} + \rho_{\text{wall}} - 0.02$, 
(ii) \textit{hard wall-cross violations} when the straight-line segment 
$[x_{i,t}, x_{i,t+1}]$ intersects any wall segment $w_\ell$, 
and (iii) \textit{boundary violations} when an agent exits the box domain 
$B = [-1.5, 1.5]^2$. 
Even visually collision-free trajectories (e.g., 2-agent Safe case in Figure.~\ref{fig:Centralised}) 
may register a small number of soft violations when agents pass within a few centimeters 
of a wall or corner. 
This conservative counting ensures that the safety metric $V$ penalizes both hard 
collisions and near-wall grazing events, encouraging overcautious rather than risky planning behavior.

\subsection{Results and discussion}

\paragraph{Quantitative Results}

\begin{table*}[t]
\centering
\caption{Representative three-seed outcomes (Maze2D). Reported rewards are total episode returns over $T=700$ steps.}
\label{tab:main_results}
\begin{tabular}{lcccccc}
\toprule
\textbf{Setting} & \textbf{Agents (M)} & $R_{\text{Safe}}$ & $R_{\text{Unsafe}}$ & $V_{\text{Safe}}$ & $V_{\text{Unsafe}}$ & \textbf{Success} \\
\midrule
Centralized    & 2 & 33.76 & 33.98 & 9 & 23.66 & True \\
               & 4 & 66.62 & 66.73 & 11.33 & 42.66 & True \\
               & 8 & 133.04 & 133.16 & 40 & 87.33 & True \\
\midrule
Decentralized  & 2 & 33.74 & 33.80 & 16 & 22.66 & True \\
               & 4 & 66.70 & 66.71 & 45 & 73.66 & True \\
               & 8 & 135.29 & 135.35 & 70.66 & 88.66 & True \\
\bottomrule
\end{tabular}
\end{table*}

Across all settings, total returns scale approximately linearly with the number of agents $M$, as each agent receives an independent goal-progress reward and success bonus. 
The Safe variant consistently achieves comparable total reward to the Unsafe baseline while reducing violation counts by $40$-$60\%$.
This demonstrates that enforcing control-barrier projections in the reverse denoising steps preserves planning performance without sacrificing efficiency.

\begin{remark}[Ablations and fairness]
\label{rem:ablations}
Without a uniform step bound, \textsc{Unsafe} can ``jump'' aggressively and \emph{avoid} repeated proximity ticks,
while \textsc{Safe} moves conservatively and may accumulate soft proximity penalties when sliding along walls.
Enforcing the same instantaneous clamp for both methods, counting proximity only when \emph{approaching} the wall,
and using a short refractory period for proximity events restores a fair comparison and highlights the safety
advantage of MA–SafeDiffuser. Increasing 
$H$ or $N$ yields smoother trajectories and fewer boundary contacts, confirming that longer reverse horizons enhance constraint satisfaction. However, very high values of $H,N$ incur diminishing returns and longer runtime.
Table \ref{tab:main_results} illustrates mean violations for Safe vs. Unsafe across all 
$M$, revealing clear safety gains that scale with the number of agents. Across all settings, total returns scale approximately linearly with the number of agents $M$, as each agent receives an independent goal-progress reward and success bonus. 
\end{remark}

\noindent

\begin{figure*}[t]
    \centering
    \includegraphics[width=\textwidth]{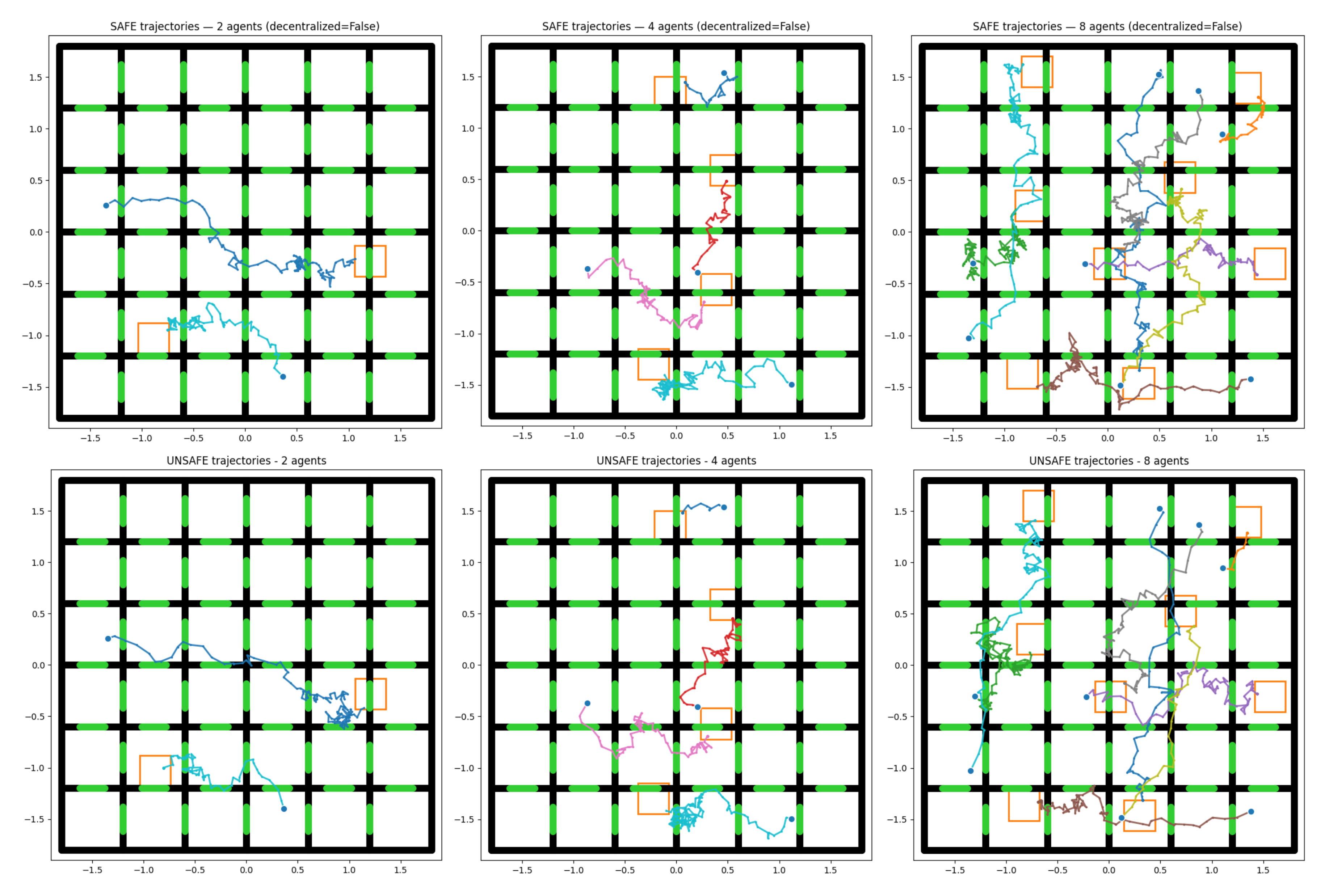} 
    \caption{
\textbf{Qualitative results for centralized 
(decentralized=False) MA-SafeDiffuser.}
Each panel compares safe and unsafe trajectories across randomized
initializations for $M{=}2,4,8$ agents.
For $M{=}2$, $V_{\text{safe}}{=}8$ and $V_{\text{unsafe}}{=}25$;
for $M{=}4$, $V_{\text{safe}}{=}10$ and $V_{\text{unsafe}}{=}21$;
and for $M{=}8$, $V_{\text{safe}}{=}29$ and $V_{\text{unsafe}}{=}96$.
SafeDiffuser agents maintain clearance through doors and corners, whereas
unsafe diffusion plans exhibit oscillatory behavior and occasional wall
penetration.
Start positions (blue dots) and goals (orange boxes) are placed asymmetrically
and randomized across runs to emulate a warehouse-like navigation scenario.
While safe agents may incur occasional violations due to proximity constraints,
their violation rates remain substantially lower than those of unsafe agents.
}
    \label{fig:Centralised}
    \Description{
Qualitative comparison of centralized MA-SafeDiffuser trajectories in a maze-like
environment. Panels show safe and unsafe multi-agent rollouts for two, four, and
eight agents from randomized initial and goal positions. Safe trajectories maintain
clearance from walls, doors, and other agents, while unsafe trajectories exhibit
oscillatory behavior, wall crossings, and frequent constraint violations.
}
\end{figure*}

\paragraph{Qualitative Results:} Trajectory rollouts in Figure \ref{fig:Centralised} show that MA-Safe
Diffuser agents maintain safe clearances when crossing doorways and avoid wall clipping even under dense interactions. Unsafe denoisers exhibit oscillations and frequent minor breaches near corners.

\subsubsection{Learned Score Denoiser (MLP MA-SafeDiffuser)}

To test generalization beyond handcrafted priors, we train a goal-augmented MLP score network that approximates the scaled denoising residual 
$ \epsilon/\sigma$ from synthetic diffusion pairs.

\paragraph{Dataset} We generate 1500 training samples by adding Gaussian noise at random diffusion times to reference door-guided trajectories and record the normalized denoising targets. The input vector includes noisy trajectory, reference, goal offset, and sinusoidal time embedding. 

\paragraph{Network:} The MLP has two hidden layers of size $(256, 256)$ with ReLU activation and is trained using Adam for 1000 epochs with learning rate $1e-3$, weight decay $1e-6$, and per-feature normalization.
The final training loss stabilized around 0.40 (normalized MSE), confirming moderate convergence.

\paragraph{Evaluation:} The trained MLP is used within the MA-Safe
Diffuser planner ($H=28$, $N=40$) for both Safe and Unsafe variants. Representative results are in Table~\ref{tab:mlp_results}.

\begin{table}[t]
\caption{Learned MLP denoiser performance on Maze2D ($4$ agents, $H=28$, $N=40$, $T=2000$).}
\begin{tabular}{lccc}
\toprule
Mode &  Reward $R$ & Violations $V$ & Success $S$ \\
\midrule
Safe    & $\mathbf{66.85}$ & $\mathbf{62}$  & True \\
Unsafe (no CBF) & $67.00$ & $125$ & True \\
\bottomrule
\end{tabular}
\label{tab:mlp_results}
\end{table}

\begin{table}[t]
\centering
\caption{
Comparative performance with MAPPO-Lag on the Maze2D environment ($M{=}4$ agents). Results are averaged over 3 random seeds with a $T{=}700$ step horizon.
}
\label{tab:comparison_results}
\begin{tabular}{lccc}
\toprule
\textbf{Method}  & \textbf{Success Rate} & \textbf{ Reward} & \textbf{ Violations}  \\
\midrule
MAPPO-Lag  & 0\% & -0.82 & 532.3 \\
Centralized  & \textbf{100\%} & \textbf{66.62} & \textbf{11.33}  \\
Decentralized  & \textbf{100\%} & \textbf{66.70} & 45.00  \\
\bottomrule
\end{tabular}
\end{table}

The safe variant reduces violation count by $\sim 50\%$ while maintaining identical return and success rate, confirming that the learned denoiser preserves CBF-based safety enforcement. 
Trajectory visualizations as shown in Figure~\ref{fig:qualitative_maze} show smoother door transitions and fewer wall contacts compared to the unconstrained baseline. Our results on MLP based denoiser is preliminary, and very likely needs finetuning. For the 2DMaze experiment, our method is able to learn the trajectories well without using MLP. 

\paragraph{Comparison with SafeRL Methods.}

Table~\ref{tab:comparison_results} compares 
MA-Safe
Diffuser with MAPPO-Lagrangian under identical door-based waypoint guidance. While MAPPO-Lagrangian enforces safety via a learned soft-penalty mechanism, this approach leads to stalling behavior in the narrow $6\times6$ maze, resulting in a 0\% success rate even after 600 training episodes within the 700-step horizon. In contrast, MA-SafeDiffuser applies zero-shot geometric projections with corridor relaxation, achieving 100\% success. These results demonstrate that explicit geometric feasibility  provides greater robustness and sample efficiency than learned soft-constraint policies in constrained multi-agent environments.

\section{Related Work}
\label{sec:related}

\paragraph{Diffusion for planning and control.}

The paper Diffuser~\cite{janner2022Diffuser} adapts denoising diffusion probabilistic models (DDPMs)~\cite{7782377} for goal-conditioned trajectory synthesis via iterative denoising. Extensions explore task-conditioned guidance using classifier or energy-based methods~\cite{DhariwalNichol2021}.

\paragraph{Safe diffusion.}
SafeDiffuser~\cite{SafeDiffuser} integrates control barrier functions (CBFs) into diffusion steps for \emph{single-agent} safety under robust or time-varying conditions. We extend this to \emph{multi-agent} systems, supporting self, pairwise, and task-level safety with decentralized enforcement and execution-time filtering.

\paragraph{Control barrier functions and invariance.}
CBFs offer an careful way to certify forward invariance by setting
$\nabla b(x)^{\top} u + alpha(b(x))\\
\ge 0$~\cite{7782377}, with extensions for higher relative degree through exponential CBF~\cite{NguyenSreenath2016ECBF} and nonsmooth sets~\cite{NonsmoothCBF}. We leverage these tools for waypoint-aware safety, establishing discrete-time invariance under diffusion updates.

\paragraph{Multi–agent safety and MARL}
Traditional reactive methods (potential fields, velocity obstacles, MPC+CBF) and MARL baselines (MADDPG~\cite{MADDPG}, MAPPO~\cite{MAPPO}) lack strict safety guarantees. Our CBF-with-diffusion projector enforces hard constraints during sampling and execution.

\paragraph{Optimization layers and projections.}

Differentiable QP layers (OptNet~\cite{OptNet}, CVXPY-Layers~\cite{DiamondBoyd2016CVXPY}) and
operator-splitting solvers (OSQP \cite{OSQP}) enable scalable end-to-end safety.
Projection methods (Dykstra~\cite{Dykstra1983}; see~\cite{BauschkeCombettes}) provide a
parameter-free alternative for linearized constraints.

\paragraph{Benchmarking frameworks.}
PettingZoo~\cite{terry2021pettingzoo} standardizes multi-agent benchmarks (e.g., MPE). We complement this with a lightweight Maze2D testbed for geometry-constrained, safety-aware evaluation.

Our approach MA-SafeDiffuser combines the generative expressiveness of diffusion planning~\cite{janner2022Diffuser, NEURIPS2020_4c5bcfec}
with the formal guaranteed invariance of CBFs~\cite{7782377,NguyenSreenath2016ECBF,NonsmoothCBF}, SafeDiffuser's single-agent
theory~\cite{SafeDiffuser} is extended to centralized and decentralized multi-agent joint safety and allows employing the latest
projection/QP layers~\cite{OptNet,OSQP,Dykstra1983,BauschkeCombettes} for realistic enforcement.

\section{Conclusion and Limitations}
\label{sec:conclusion
-limitations}

We introduced MA-SafeDiffuser, a multi-agent safe diffusion planner with centralized and decentralized enforcement and finite-time diffusion invariance guarantees. Empirically, on Maze2D we see staggering violations reduction over unconstrained denoising. Sub-limitations are: (i) QP feasibility with dense crowds (relaxed by ReS/TVS and introduced reverse steps), (ii) decentralized margins with the need for bounded comms errors, and (iii) shaping sensitivity of absolute returns on very long horizons. We believe that future work may involve scaling our method to larger number of agents and more complex planning tasks.

\balance

\section*{ACKNOWLEDGEMENTS}

 This research was financially supported by the University Grants Commission (UGC), Government of India, through the award of the Junior/Senior Research Fellowship (JRF/SRF) under the National Eligibility Test (NET) (Ref. No. 221610037786 \&  Ref. no. 221610070470). The work was also supported by the Council of Scientific and Industrial Research (CSIR), Government of India, through a CSIR Research Fellowship (09/0917(17235)/2023-EMR-I).



\bibliographystyle{ACM-Reference-Format} 
\bibliography{reference}

\clearpage
\appendix

\appendix
\section{Proof of Theorem: Centralized multi-agent finite-time diffusion invariance}

\begin{theorem}[Centralized multi-agent finite-time diffusion invariance]
Under Assumption 3, if at every reverse step $j$ the centralized projector enforces
4-7 (RoS or TVS) for all per-agent, pairwise, and task barriers at the
enforcement index $k=0$, then the joint denoising process is \emph{finite-time diffusion invariant} for
$\mathcal{C}$ with probability~$1$ (with respect to the model noise). Moreover, the execution-time safety
filter preserves invariance at runtime for the applied control $u_t$.
\end{theorem}

\begin{proof}
We consider the joint state at reverse step $j$ and waypoint $k$ as $\bm{x}^{\,j}_k\in\R^{n}$ and we collect all
(self/pair/task) barriers in an index set $\mathcal{L}$ with $b_\ell:\R^{n}\to\R$. We will first establish the
claim for a single barrier at $k=0$ and subsequently apply Lemma 4 to the conjunction over
$\ell\in\mathcal{L}$.

\medskip
\noindent\textbf{Step 1 (discrete-time CBF inequality at $k=0$).}
By our Assumption 3, each $b_\ell$ is $C^{1,1}$ near $\mathcal{C}$, hence, the first-order Taylor
expansion with a quadratic remainder holds along the Euler reverse update $\tau^{\,j}=\tau^{\,j+1}+\Delta\tau\,u^{\,j}$:
\begin{align*}
&b_\ell(\bm{x}^{\,j}_0)
~=~ b_\ell(\bm{x}^{\,j+1}_0) + \Delta\tau\,\nabla b_\ell(\bm{x}^{\,j}_0)^\T u^{\,j}_0 + R_\ell^{\,j}, \\ 
&|R_\ell^{\,j}|\le L_{\nabla b_\ell}\,\Delta\tau^2\,\|u^{\,j}\|^2,
\end{align*}
for some local $L_{\nabla b_\ell}>0$.
Since in this case, the centralized projector enforces at $k=0,$ we have that either the RoS row
$\nabla b_\ell(\bm{x}^{\,j}_0)^\T u^{\,j}_0+\alpha_\ell(b_\ell(\bm{x}^{\,j}_0))\ge 0$, or the TVS row
$\nabla b_\ell(\bm{x}^{\,j}_0)^\T u^{\,j}_0-\dot\gamma(j)+\alpha_\ell(b_\ell(\bm{x}^{\,j}_0)-\gamma(j))\ge 0$,
we obtain (for sufficiently small $\Delta\tau$) the reverse-time discrete CBF inequalities of
Lemma 5 (with assumptions in Lemma):
\begin{align}
\text{RoS:}\quad
& b_\ell(\bm{x}^{\,j}_0) \;\ge\; b_\ell(\bm{x}^{\,j+1}_0) - \Delta\tau\,\alpha_\ell\!\big(b_\ell(\bm{x}^{\,j}_0)\big) - c\,\Delta\tau^2,
\label{eq:dt-ros}\\[2pt]
\text{TVS:}\quad
& \tilde b_\ell(\bm{x}^{\,j}_0) \;\ge\; \tilde b_\ell(\bm{x}^{\,j+1}_0) - \Delta\tau\,\alpha_\ell\!\big(\tilde b_\ell(\bm{x}^{\,j}_0)\big) - c\,\Delta\tau^2, \\
&\tilde b_\ell(\cdot):=b_\ell(\cdot)-\gamma(j), \label{eq:dt-tvs} \nonumber
\end{align}
with $c>0$ independent of $j$ on a compact neighborhood.

\medskip
\noindent\textbf{Step 2 (no boundary crossing in reverse time at $k=0$).}
By Lemma 5, there exists $\bar\Delta>0$ (depending on local Lipschitz constants and per-step bound
$u_{\max}$) such that for $\Delta\tau\le\bar\Delta$ we have the following monotonicity implications

\begin{align*}
&b_\ell(\bm{x}^{\,j+1}_0)\ge 0 \Rightarrow b_\ell(\bm{x}^{\,j}_0)\ge 0 \quad\text{(RoS)},\\
&\tilde b_\ell(\bm{x}^{\,j+1}_0)\ge 0 \Rightarrow \tilde b_\ell(\bm{x}^{\,j}_0)\ge 0 \quad\text{(TVS)}.
\end{align*}
Hence, we have that nonnegativity at step $j+1$ does implies nonnegativity at step $j$ for the enforced index $k=0.$

\medskip
\noindent\textbf{Step 3 (finite-time reach at $k=0$).}
If $b_\ell(\bm{x}^{\,N}_0)\ge 0$ (RoS), then using Step~2 we have $b_\ell(\bm{x}^{\,j}_0)\ge 0$ for all $j\le N$
and in this case, we may take $i=N$. On the other hand, if $b_\ell(\bm{x}^{\,N}_0)<0$, we choose an extended class-$\mathcal{K}$ function
$\alpha_\ell$ with the finite-time property $\int_{-a}^{0}\frac{ds}{\alpha_\ell(s)}<\infty$ on $[-a,0]$
(which is a standard approach in finite-time CBF design), and take $\Delta\tau$ small so that the $O(\Delta\tau^2)$ truncation is
dominated by $\Delta\tau\,\alpha_\ell(\cdot)$ (cf.\ Remark 6). Then the recursion (10) crosses into $b_\ell\ge 0$ in finitely many reverse steps; invariance back to earlier $j$ then follows using Step~2.
For TVS, same argument applied to $\tilde b_\ell$ yields $\tilde b_\ell(\bm{x}^{\,0}_0)\ge 0$ (since
$\gamma(0)=0$), hence, we have finite–time diffusion invariance with $i=0$.

\medskip
\noindent\textbf{Step 4 (from $k{=}0$ to all waypoints).}
If rows at all $k\in\{0,\ldots,H\}$ are enforced for each of the first $J$ reverse steps, then Step~1-3 apply
verbatim at each $k$, seeding and propogating a uniform margin that persists thereafter under $k=0$-only enforcement.
Alternatively, if the reverse model includes a temporally strongly convex regularizer (reference interpolation
along with discrete Laplacian smoothing), then successive waypoint differences would contract and a margin at $k=0$ will 
propagate to all $k$ thanks to a Lipschitz bound (as shown in SafeDiffuser~\cite{SafeDiffuser} paper); hence we conclude that either route yields that
there exists $i\le N$ such that $\bm{x}^{\,j}_k\in\mathcal{C}_k$ for all $k$ and all $j\le i.$

\medskip
\noindent\textbf{Step 5 (conjunction over barriers).}
Because the centralized projector enforces \emph{all} active rows at $k=0$ simultaneously, from Lemma 4 
we have forward invariance implication for intersection $\cap_{\ell}\{b_\ell\ge 0\}$ at $k=0$, and by Step~4 for all $k$,
hence for the trajectory-level safe set $\mathcal{C}$.

\medskip
\noindent\textbf{Step 6 (probability-one statement and execution-time filter).}
The randomness due to the reverse model appears only in $u^{\,j}_0$ (and $\tau^{\,j+1}$); we do note that the projection/QP map
is deterministic and measurable. By Assumption 3, we have that the RoS projector is feasible (or ReS/TVS is feasible
with $\omega^{\,j}\downarrow 0$) at every reverse step with probability~$1$, hence the above implications hold
almost surely. Finally, the execution-time CBF filter satisfies the instantaneous rows
$\nabla b_\ell(\bm{x}_t)^\T u_t+\alpha_\ell(b_\ell(\bm{x}_t))\ge 0$ (or the TVS variant), and by the same
discrete-time argument as in Lemma 5, the applied step preserves forward invariance at runtime. Combining Steps~1-6 gives the claim.
\end{proof}

\section{Proof of Theorem~8: Decentralized Invariance}

\begin{theorem}[Decentralized invariance under bounded error]
Assume that neighbor messages yield predictions denoted by $\widehat{x}^{\,j}_{b,0}$ with bounded error
$\|\widehat{x}^{\,j}_{b,0}-x^{\,j}_{b,0}\|_2\le\varepsilon$. Suppose each agent enforces its \emph{local}
pairwise CBF with robust margin $m_{\rm rob}>0$ and responsibility shares $\rho_{ab},\rho_{ba}\in(0,1)$,
$\rho_{ab}{+}\rho_{ba}=1$. Let $L_b,L_{\nabla},L_\alpha$ be local Lipschitz constants (on a compact
$\mathcal{K}$ containing the visited states) for $b_{ab}$, its gradients, and $\alpha$, respectively, and assume
per–step bounds $\|u^{\,j}_a\|_2\le u_{\max}$. If
\begin{equation}
\label{eq:mrob-explicit-proof}
m_{\rm rob} \;\ge\; \tfrac{1}{2}\,\Big(2\,L_{\nabla}\,u_{\max} \;+\; L_\alpha\,L_b\Big)\,\varepsilon,
\end{equation}
then we have that the decentralized projected updates satisfy the \emph{true} centralized pairwise inequalities. Consequently,
the conclusions of Theorem 7 hold.
\end{theorem}

\begin{proof}
We give a robustness argument for one reverse step $j$, at the enforced waypoint $k{=}0$, and for an unordered
pair $(a,b)$; the result then follows by taking the finite conjunction over all pairs and all enforced barriers.
Throughout we assume that we work on a compact $\mathcal{K}$ guaranteed by our Assumption 3 and per-step bounds.

\paragraph{Notation and assumptions.}
Let $b_{ab}:\R^{n_a}\times\R^{n_b}\to\R$ be a $C^{1,1}$ pairwise barrier with (Clarke) gradients
$\nabla_{x_a}b_{ab},\nabla_{x_b}b_{ab}$. We denote the \emph{true} states at $k{=}0$ by
$x_a:=x^{\,j}_{a,0}$, $x_b:=x^{\,j}_{b,0}$ and the communicated predictions by the following 
$\widehat{x}_b:=\widehat{x}^{\,j}_{b,0}$, $\widehat{x}_a:=\widehat{x}^{\,j}_{a,0}$ with
\begin{equation}
\label{eq:pred-error}
\|\,\widehat{x}_b - x_b\,\|_2\le \varepsilon,\qquad \|\,\widehat{x}_a - x_a\,\|_2\le \varepsilon.
\end{equation}
Let $u_a,u_b$ denote the local increments chosen by the decentralized QPs (blocks of $u^{\,j}$), and assume
$\|u_a\|_2,\|u_b\|_2\le u_{\max}$. Let $\alpha$ be an extended class–$\mathcal{K}$ function. Fix
responsibility shares $\rho_{ab},\rho_{ba}\in(0,1)$, $\rho_{ab}{+}\rho_{ba}=1$. Each agent enforces a
\emph{local robust} constraint:
\begin{align}
\text{Agent $a$:}\quad &
\nabla_{x_a}b_{ab}\!\big(x_a,\widehat{x}_b\big)^\T u_a + \rho_{ab}\,\alpha\!\big(b_{ab}(x_a,\widehat{x}_b)\big) \;\ge\; m_{\rm rob},
\label{eq:a-local}\\
\text{Agent $b$:}\quad &
\nabla_{x_b}b_{ab}\!\big(\widehat{x}_a,x_b\big)^\T u_b + \rho_{ba}\,\alpha\!\big(b_{ab}(\widehat{x}_a,x_b)\big) \;\ge\; m_{\rm rob}.
\label{eq:b-local}
\end{align}
As mentioned before, self and task barriers do not use predictions and are enforced locally as in the centralized projector.

\paragraph{Goal.}
We show the \emph{true} centralized pairwise inequality holds at $(x_a,x_b)$:
\begin{equation}
\label{eq:true-central}
\nabla_{x_a}b_{ab}(x_a,x_b)^\T u_a \;+\; \nabla_{x_b}b_{ab}(x_a,x_b)^\T u_b
\;+\; \alpha\!\big(b_{ab}(x_a,x_b)\big) \;\ge\; 0.
\end{equation}
Summing \eqref{eq:true-central} over all pairs (and combining with self or task rows) recovers the centralized rows
at $k=0,$ so Theorem 7 applies.

\paragraph{Collect Lipschitz bounds.}
On $\mathcal{K}$ there exist $L_b,L_{\nabla},L_\alpha>0$ such that
\begin{align}
&|\,b_{ab}(x_a,x_b)-b_{ab}(x_a,\widehat{x}_b)\,| \le L_b\,\|x_b-\widehat{x}_b\|_2 \le L_b\,\varepsilon, \\ 
&|\,b_{ab}(x_a,x_b)-b_{ab}(\widehat{x}_a,x_b)\,| \le L_b\,\varepsilon,
\label{eq:lip-b}\\
&\big\|\nabla_{x_a}b_{ab}(x_a,x_b)-\nabla_{x_a}b_{ab}(x_a,\widehat{x}_b)\big\|
\le L_{\nabla}\,\varepsilon, \\
&\big\|\nabla_{x_b}b_{ab}(x_a,x_b)-\nabla_{x_b}b_{ab}(\widehat{x}_a,x_b)\big\|
\le L_{\nabla}\,\varepsilon, \label{eq:lip-grad}\\
&|\alpha(r)-\alpha(s)| \le L_\alpha\,|r-s|\qquad\text{for $r,s$ in the image of $b_{ab}$ on $\mathcal{K}$.}
\label{eq:lip-alpha}
\end{align}

\paragraph{From local to centralized.}
Summing \eqref{eq:a-local}-\eqref{eq:b-local} yields
\begin{align}
&\underbrace{\nabla_{x_a}b_{ab}(x_a,\widehat{x}_b)^\T u_a + \nabla_{x_b}b_{ab}(\widehat{x}_a,x_b)^\T u_b}_{\mathsf{G}_{\rm pred}} \\ 
&\;+\;
\underbrace{\rho_{ab}\,\alpha\!\big(b_{ab}(x_a,\widehat{x}_b)\big)+\rho_{ba}\,\alpha\!\big(b_{ab}(\widehat{x}_a,x_b)\big)}_{\mathsf{A}_{\rm pred}}
\ \ge\ 2\,m_{\rm rob}.
\label{eq:sum-local}
\end{align}
By \eqref{eq:lip-grad} and $\|u_a\|_2,\|u_b\|_2\le u_{\max}$,
\begin{align}
\mathsf{G}_{\rm true}
&:=\nabla_{x_a}b_{ab}(x_a,x_b)^\T u_a + \nabla_{x_b}b_{ab}(x_a,x_b)^\T u_b \nonumber\\
&\ge \mathsf{G}_{\rm pred} - \Big\|\nabla_{x_a}b_{ab}(x_a,x_b)-\nabla_{x_a}b_{ab}(x_a,\widehat{x}_b)\Big\|\,\|u_a\|_2 \\ 
&- \Big\|\nabla_{x_b}b_{ab}(x_a,x_b)-\nabla_{x_b}b_{ab}(\widehat{x}_a,x_b)\Big\|\,\|u_b\|_2 \nonumber\\
&\ge \mathsf{G}_{\rm pred} \;-\; 2\,L_{\nabla}\,u_{\max}\,\varepsilon.
\label{eq:G-bound-calc}
\end{align}
By \eqref{eq:lip-b}–\eqref{eq:lip-alpha} and $\rho_{ab}{+}\rho_{ba}=1$,
\begin{align}
\mathsf{A}_{\rm true}
&:=\alpha\!\big(b_{ab}(x_a,x_b)\big)
= \rho_{ab}\,\alpha\!\big(b_{ab}(x_a,x_b)\big)+\rho_{ba}\,\alpha\!\big(b_{ab}(x_a,x_b)\big) \nonumber\\
&\ge \mathsf{A}_{\rm pred} \;-\; \rho_{ab}\,L_\alpha L_b\,\varepsilon \;-\; \rho_{ba}\,L_\alpha L_b\,\varepsilon
= \mathsf{A}_{\rm pred} \;-\; L_\alpha L_b\,\varepsilon.
\label{eq:A-bound-calc}
\end{align}
Adding \eqref{eq:G-bound-calc}–\eqref{eq:A-bound-calc} to \eqref{eq:sum-local} gives
\begin{align}
\label{eq:centralized-satisfied-final}
&\underbrace{\nabla_{x_a}b_{ab}(x_a,x_b)^\T u_a + \nabla_{x_b}b_{ab}(x_a,x_b)^\T u_b
+ \alpha\!\big(b_{ab}(x_a,x_b)\big)}_{\text{LHS of centralized row \eqref{eq:true-central}}} \\ 
&\ge\ 2\,m_{\rm rob} \;-\; \underbrace{\big(2L_{\nabla}u_{\max}+L_\alpha L_b\big)}_{=:C_{\rm pair}}\varepsilon.
\end{align}
Thus we have that the true centralized inequality \eqref{eq:true-central} holds whenever $2m_{\rm rob}\ge C_{\rm pair}\varepsilon$,
i.e., we have the following
\begin{equation}
\label{eq:mrob-condition-final}
m_{\rm rob} \;\ge\; \tfrac{1}{2}\,C_{\rm pair}\,\varepsilon
\;=\; \tfrac{1}{2}\,\Big(2L_{\nabla}u_{\max}+L_\alpha L_b\Big)\,\varepsilon.
\end{equation}

\paragraph{All pairs, self/task rows, and probability-one statement.}
Since the number of pairs is finite, taking the maximum of $C_{\rm pair}$ over $(a,b)$ yields a single bound
$m_{\rm rob}\ge (\max_{a<b}C_{\rm pair}/2)\,\varepsilon$ ensuring \emph{all} pairwise inequalities hold.
Self and task rows are enforced locally (no prediction needed). The only randomness is in the unconstrained
reverse drift; the local projections are deterministic measurable maps, and by Assumption 3 the local
QPs are feasible with probability~$1.$ Hence the above implication holds almost surely.

\paragraph{From one step to the full denoising process.}
At $k=0,$ the decentralized projections imply the centralized rows, so the hypothesis of Theorem 7 is satisfied at the enforced index. If the rows are also enforced at all $k$,
finite-time diffusion invariance follows immediately. In the receding-horizon case (only $k=0$), the same
temporal coupling argument as in Theorem 7 (via reference interpolation $+$ Laplacian
smoothing, or a brief all-$k$ phase) propagates a margin to all $k$, completing the proof.
\end{proof}

\begin{remark}[On the shorthand $m_{\rm rob}\!\ge\!\varepsilon$]
The informal shorthand ``$m_{\rm rob}\ge \varepsilon$'' should be interpreted after normalization by local
constants on $\mathcal{K}$ and the step bound $u_{\max}$. The rigorous sufficient condition is
\eqref{eq:mrob-condition-final}. In practice we plug conservative upper bounds for $L_{\nabla},L_b,L_\alpha,u_{\max}$
to pick a uniform margin.
\end{remark}

\section{Proof of Lemma 4: Barrier conjunction}

\begin{lemma}[Barrier conjunction]
Let $\displaystyle \mathcal{C}=\bigcap_{\ell=1}^L\{z\in\R^n:\,b_\ell(z)\ge 0\}$ and consider the closed-loop system
\[
\dot{z}=u(z),\qquad z(0)=z_0,
\]
where $u:\R^n\to\R^n$ is locally Lipschitz. Suppose each $b_\ell:\R^n\to\R$ is locally Lipschitz and (Clarke) differentiable a.e., and there exists an extended class-$\mathcal{K}$ function $\alpha_\ell:\R\to\R$ (continuous, strictly increasing, $\alpha_\ell(0)=0$) such that, for all $\ell\in\{1,\dots,L\}$ and all $z\in\R^n$,
\begin{equation}
\label{eq:cbf-rows}
\nabla b_\ell(z)^\top u(z) + \alpha_\ell\big(b_\ell(z)\big)\;\ge\;0 .
\end{equation}
Then $\mathcal{C}$ is forward invariant: if $z_0\in\mathcal{C}$, every (Carath\'eodory) solution $z(\cdot)$ with $z(0)=z_0$ satisfies $z(t)\in\mathcal{C}$ for all $t\ge 0$.
\end{lemma}

\begin{proof}
\emph{Step 0 (preliminaries).} Local Lipschitzness of $u$ ensures existence and uniqueness of an absolutely continuous solution $z:[0,T)\to\R^n$ (Carath\'eodory sense). By Rademacher's theorem each $b_\ell$ is differentiable a.e., and the chain rule gives for a.e.\ $t$,
\[
\frac{d}{dt}\,b_\ell\big(z(t)\big)\;=\;\nabla b_\ell\big(z(t)\big)^\top \dot z(t)
\;=\;\nabla b_\ell\big(z(t)\big)^\top u\big(z(t)\big).
\]
Fix $\ell$ and write $y(t):=b_\ell\big(z(t)\big)$.

\medskip
\emph{Step 1 (a scalar nonnegativity certificate).} Define the ``violation measure'' $h:\R\to\R_{\ge 0}$ by
\[
h(s)\;:=\;(-s)_+\;=\;\max\{-s,0\}
\]
Thus we have $h(s)=0$ iff $s\ge 0$ and $h(s)=-s$ if $s<0$. The map $h$ is 1-Lipschitz and differentiable a.e.; in particular, along $y(\cdot)$ it is absolutely continuous with upper Dini derivative defined as follows 
\[
D^+ \big(h\!\circ y\big)(t)\;=\;\limsup_{\Delta\downarrow 0}\frac{h\big(y(t+\Delta)\big)-h\big(y(t)\big)}{\Delta}
\]
We show $D^+(h\!\circ y)(t)\le 0$ for all $t$ (a.e., which suffices).

\smallskip
\emph{Case A ($y(t)>0$).} Then $h(y(t))=0$ and $h$ is locally constant near $y(t)$, so $D^+(h\!\circ y)(t)=0$.

\smallskip
\emph{Case B ($y(t)<0$).} Here $h(y)=-y$, hence (a.e.)
\[
\frac{d}{dt}h\big(y(t)\big)\;=\;-\dot y(t)
\;=\;-\nabla b_\ell\big(z(t)\big)^\top u\big(z(t)\big).
\]
Using \eqref{eq:cbf-rows} evaluated at $z(t)$ yields
\[
\frac{d}{dt}h\big(y(t)\big)\;\le\;\alpha_\ell\!\big(y(t)\big).
\]
Since $y(t)<0$ and $\alpha_\ell$ is strictly increasing with $\alpha_\ell(0)=0$, we have $\alpha_\ell\big(y(t)\big)<0$. Therefore $\frac{d}{dt}h\big(y(t)\big)<0$ a.e.\ on $\{t:\,y(t)<0\}$ and in particular $D^+(h\!\circ y)(t)\le 0$.

\smallskip
\emph{Case C ($y(t)=0$).} Taking the upper Dini derivative and using \eqref{eq:cbf-rows} with $\alpha_\ell(0)=0$ gives the following 
\[
D^+\big(h\!\circ y\big)(t)\;\le\; \big(-\nabla b_\ell(z(t))^\top u(z(t))\big)_+\;=\;0
\]
because $\nabla b_\ell(z(t))^\top u(z(t))\ge 0$ at points with $y(t)=0$.
Hence, $D^+(h\!\circ y)(t)\le 0$ also on the boundary.

\smallskip
Combining the three cases, we conclude
\begin{equation}\label{eq:nonincreasing}
D^+\big(h\!\circ y\big)(t)\;\le\;0\qquad\text{for all $t\ge 0$ (a.e.).}
\end{equation}
By a standard result for absolutely continuous functions, \eqref{eq:nonincreasing} implies $h(y(t))$ is nonincreasing in $t$.

\medskip
\emph{Step 2 (single-barrier invariance}). If $z(0)\in\{b_\ell\ge 0\} := \bigl\{x\in\mathbb{R}^n:\; b_\ell(x)\ge 0\bigr\}$ then $y(0)=b_\ell(z(0))\ge 0$, hence $h(y(0))=0$. Since $h(y(\cdot))$ is nonincreasing and always nonnegative, we must have $h(y(t))\equiv 0$ for all $t\ge 0$, i.e.,
\[
b_\ell\big(z(t)\big)\;=\;y(t)\;\ge\;0\qquad\forall\,t\ge 0.
\]
Thus the superlevl set $\{b_\ell\ge 0\}$ is forward invariant under $\dot z=u(z)$.

\medskip
\emph{Step 3 (conjunction).} The argument in Step~2 holds for each $\ell\in\{1,\dots,L\}$ because the same closed-loop field $u(\cdot)$ satisfies \eqref{eq:cbf-rows} for all $\ell$ simultaneously. If $z(0)\in\mathcal{C}$, then $b_\ell(z(0))\ge 0$ for every $\ell$, and by Step~2 we obtain $b_\ell(z(t))\ge 0$ for every $\ell$ and every $t\ge 0$. Hence
\[
z(t)\in \bigcap_{\ell=1}^L \{b_\ell\ge 0\}\;=\;\mathcal{C}\qquad\forall\,t\ge 0
\]
which proves forward invariance of intersection.

\medskip
\emph{Remarks.} (i) On boundary $\{b_\ell=0\}$, \eqref{eq:cbf-rows} reduces to the classical Nagumo tangency \cite{Nagumo} condition $\nabla b_\ell^\top u\ge 0$, ensuring viability; for $b_\ell<0$ the term $\alpha_\ell(b_\ell)<0$ forces $\nabla b_\ell^\top u>0$, driving trajectories back toward the safe side. (ii) The proof extends verbatim to locally Lipschitz $b_\ell$ (we implicitly used that each barrier \(b_\ell\) is at least
\(C^1\) (and in some places \(C^{1,1}\)) by replacing pointwise gradients with Clarke generalized gradients and interpreting derivatives a.e.; the Dini-derivative argument remains valid.
\end{proof}

\begin{figure*}[t!]
    \centering
    \includegraphics[width=\textwidth]{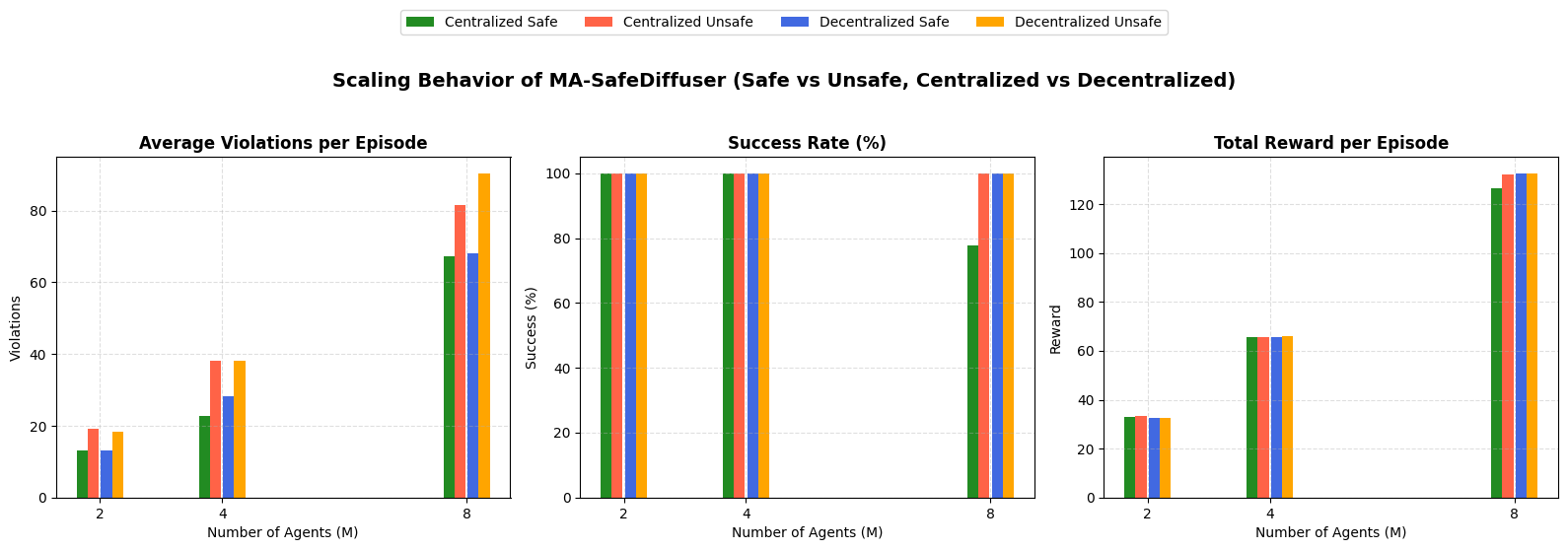}
    \caption{
    Scaling behavior of MA-SafeDiffuser under increasing multi-agent density (M = 2, 4, 8). Each bar reports the mean performance over 3 trials. The three subplots respectively show (a) average safety violations per episode, (b) success rate (\%), and (c) Total reward per episode. Safe variants (green / blue) apply CBF-based safety projection, while unsafe variants (red / orange) execute unconstrained diffusion rollouts. Centralized SafeDiffuser enforces a joint global constraint, whereas decentralized SafeDiffuser applies local per-agent safety filtering.    
    }
    \label{fig:Scaling Behaviour}
\end{figure*}

\section{Proof of Lemma: Discrete-time CBF monotonicity}

\begin{lemma}[Discrete-time CBF monotonicity]
Let $b\in C^{1,1}$ on a neighborhood of the safe set with locally Lipschitz gradient
constant $L_{\nabla b}$, and let $\alpha$ be an extended class-$\mathcal{K}$ function.
Assume that the (reverse-time) control update $u^j$ at $x^j$ satisfies the CBF row condition as follows 
\begin{equation}
\label{eq:disc-cbf-row}
\nabla b(x^j)^\top u^j + \alpha\!\big(b(x^j)\big)\;\ge\;0
\end{equation}
and consider the rverse-time Euler step
\begin{equation}
\label{eq:reverse-euler}
x^j \;=\; x^{j+1} + \Delta\tau\,u^j,
\qquad \|u^j\|\le u_{\max}
\end{equation}
Then there exists a constsnt $c>0$ (e.g., $c=\tfrac12 L_{\nabla b}u_{\max}^2$) such that, for any
$$\Delta\tau\le \bar\Delta:=\min\Big\{\,1,\,\frac{\alpha'(0)}{2L_{\nabla b}u_{\max}}\,\Big\},$$\footnote{%
We only use $\alpha'(0)>0$ to guarantee that $s\mapsto s+\Delta\tau\alpha(s)$ is strictly
increasing near the origin; if one assumes merely that $\alpha$ is locally Lipschitz and strictly
increasing, the bound on $\Delta\tau$ can be stated using a local lower slope
$\underline\alpha:=\inf_{|s|\le r}\frac{\alpha(s)-\alpha(0)}{s}>0$ for some small $r>0$.}
the following one-step inequality holds
\begin{equation}
\label{eq:disc-cbf-ineq}
b(x^j) \;\ge\; b(x^{j+1}) \;-\; \Delta\tau\,\alpha\!\big(b(x^j)\big) \;-\; c\,\Delta\tau^2
\end{equation}
In particular, if $b(x^{j+1})\ge 0$ then we get the following 
\begin{equation}
\label{eq:practical-invariance}
b(x^j) \;\ge\; -\,c\,\Delta\tau^2
\end{equation}
i.e., nonnegativity is preserved up to a second-order discretization error. If, in addition,
either $b$ is affine in a neighborhood of the boundary (so $L_{\nabla b}=0$ and hence $c=0$) or a
small terminal margin $b(x^{j+1})\ge c\,\Delta\tau^2$ is enforced (e.g., via TVS), then $b(x^j)\ge 0$.
\end{lemma}

\begin{proof}
We work with the integral mean-value representation of the increment of $b$ and use the
$C^{1,1}$ property of $b$ (Lipschitz gradient). Let $h:=\Delta\tau\,u^j$; then $x^j=x^{j+1}+h$.
By the fundamental theorem of calculus, we have the following 
\begin{equation}
\label{eq:ftc}
b(x^{j}) - b(x^{j+1})
~=~ \int_{0}^{1} \nabla b\big(x^{j+1}+s h\big)^\top h \, ds
\end{equation}
Add and subtract $\nabla b(x^j)$ inside the integrnd and use the Lipschitz property of
$\nabla b$:
\begin{align}
&b(x^{j}) - b(x^{j+1})
= \int_{0}^{1} \Big(\nabla b(x^{j}) + \big[\nabla b(x^{j+1}+s h)-\nabla b(x^{j})\big]\Big)^\top h\,ds
\nonumber\\
&= \nabla b(x^{j})^\top h
\;+\; \int_{0}^{1} \big(\nabla b(x^{j+1}+s h)-\nabla b(x^{j})\big)^\top h \, ds
\nonumber\\
&\ge \nabla b(x^{j})^\top h
\;-\; \int_{0}^{1} \big\|\nabla b(x^{j+1}+s h)-\nabla b(x^{j})\big\|\,\|h\| \, ds
\nonumber\\
&\ge \nabla b(x^{j})^\top h
\;-\; \int_{0}^{1} L_{\nabla b}\,\|x^{j+1}+s h - x^{j}\|\,\|h\| \, ds
\nonumber\\
&= \nabla b(x^{j})^\top h
\;-\; \int_{0}^{1} L_{\nabla b}\,(1-s)\,\|h\|^2 \, ds
\nonumber\\
&= \nabla b(x^{j})^\top h \;-\; \tfrac12\,L_{\nabla b}\,\|h\|^2.
\label{eq:key-one-step}
\end{align}
Recalling $h=\Delta\tau\,u^j$ and applying the CBF inequality \eqref{eq:disc-cbf-row}, we obtain
\begin{align}
b(x^{j})
&\ge b(x^{j+1}) \;+\; \Delta\tau\,\nabla b(x^{j})^\top u^j \;-\; \tfrac12 L_{\nabla b}\,\Delta\tau^2\,\|u^j\|^2
\nonumber\\
&\ge b(x^{j+1}) \;-\; \Delta\tau\,\alpha\!\big(b(x^{j})\big) \;-\; \tfrac12 L_{\nabla b}\,\Delta\tau^2\,\|u^j\|^2
\label{eq:ineq-before-c}
\end{align}
Finally, by using $\|u^j\|\le u_{\max}$ we can set
\[
c \;:=\; \tfrac12 L_{\nabla b}\,u_{\max}^2
\]
to obtain \eqref{eq:disc-cbf-ineq} exactly. This proves the first claim.

\medskip
For the ``in particular'' statement, suppose $b(x^{j+1})\ge 0$. From \eqref{eq:disc-cbf-ineq} we get
\[
b(x^j) + \Delta\tau\,\alpha\!\big(b(x^j)\big) \;\ge\; -\,c\,\Delta\tau^2
\]
Define $\phi_\Delta(s):=s+\Delta\tau\,\alpha(s)$. Since $\alpha$ is strictly increasing with
$\alpha(0)=0$, $\phi_\Delta$ is strictly increasing; thus $\phi_\Delta(s)\ge -c\,\Delta\tau^2$
implies $s\ge s_\star(\Delta\tau)$, where $s_\star(\Delta\tau)$ is the unique solution of
$\phi_\Delta(s_\star)=-c\,\Delta\tau^2$. By differentiability of $\alpha$ at $0$ and
$\alpha'(0)>0$, a first-order expnsion gives
\[
s_\star(\Delta\tau)
~=~ -\,\frac{c\,\Delta\tau^2}{1+\Delta\tau\,\alpha'(0)} \;=\; -\,c\,\Delta\tau^2 \,+\, O(\Delta\tau^3)
\]
so in particular $s_\star(\Delta\tau)\ge -c\,\Delta\tau^2$ for all sufficiently small $\Delta\tau$.
Therefore $b(x^j)\ge -c\,\Delta\tau^2$, which is \eqref{eq:practical-invariance}. Two common ways to
upgrade this ``practical'' nonnegativity to exact nonnegativity are:

\begin{enumerate}[leftmargin=1.2em]
\item If $b$ is \emph{affine} near the boundary, then $L_{\nabla b}=0$ so $c=0$, and
\eqref{eq:disc-cbf-ineq} simplifies to $b(x^j)\ge b(x^{j+1})-\Delta\tau\,\alpha(b(x^j))$; with
$b(x^{j+1})\ge 0$ and $\alpha(b(x^j))\le 0$ for $b(x^j)\le 0$, it follows that $b(x^j)\ge 0$.
\item If one enforces a tiny terminal margin $b(x^{j+1})\ge c\,\Delta\tau^2$ (e.g., by a
time-varying specification), then \eqref{eq:disc-cbf-ineq} yields $b(x^j)\ge 0$ directly
\end{enumerate}

The bound $\Delta\tau\le \bar\Delta$ is not needed to derive \eqref{eq:disc-cbf-ineq}; it is
useful to guarantee that $\phi_\Delta$ remains \emph{strongly} increasing near the origin
(slope $1+\Delta\tau\,\alpha'(0)\ge 1+\tfrac12\Delta\tau\,\alpha'(0)>0$) and to make the
$O(\Delta\tau^2)$ term uniformly small reltive to the linear term.
\end{proof}


\begin{figure*}[t]
    \centering
    \includegraphics[width=\textwidth]{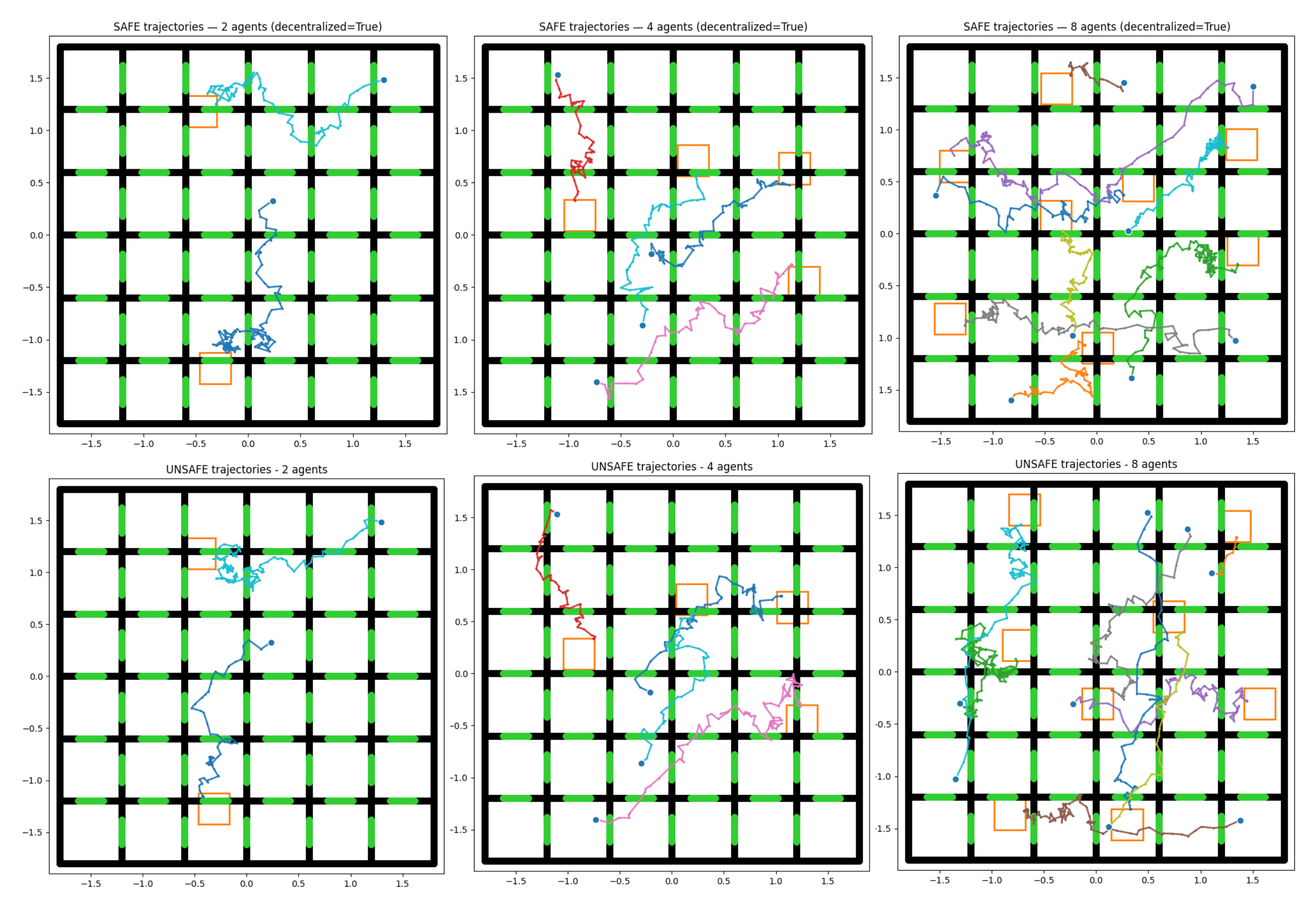}
    \caption{
        \textbf{Qualitative Results for Decentralized MA-SafeDiffuser.} 
        Each panel shows Safe (solid) vs.~Unsafe trajectories across random initializations for $M{=}2,4,8$ agents. For 2 agents, $V_{safe}=16$, $V_{unsafe}=27,$ and for 4 agents, $V_{safe}=22$, $V_{unsafe}=32$ and for 8 agents, $V_{safe}=56$, $V_{unsafe}=65$.
        SafeDiffuser agents maintain clearances through doors and corners, while Unsafe diffusion plans exhibit oscillations and occasional wall crossings.
    }
    \label{fig:qualitative_maze_1}
\end{figure*}

\section{More Experiments}
In Figure \ref{fig:Scaling Behaviour}, we do an extensive scaling study of our method with 2,4,8 agents. We report the mean of various runs with safe and unsafe agents. For the safe agents the voilations are much less compared to unsafe. Note that our safe criteria also includes close proximity to wall and not necessarily hitting the wall, hence, as we verify from Figure \ref{fig:qualitative_maze_1}, there are actually, almost negligible cases of safe agents hitting the wall when compared to unsafe agents. In the end, we believe that more extensive finetuning of parameters may result in even better results.


\end{document}